\documentclass{article}

\usepackage{amsmath,amsfonts,bm}

\def\eqref#1{equation~\ref{#1}}

\def\1{\bm{1}}

\DeclareMathAlphabet{\mathsfit}{\encodingdefault}{\sfdefault}{m}{sl}
\SetMathAlphabet{\mathsfit}{bold}{\encodingdefault}{\sfdefault}{bx}{n}

\ifx\theorem\undefined
\newtheorem{theorem}{Theorem}[section]
\fi

\ifx\example\undefined
\newtheorem{example}{Example}[section]
\fi

\ifx\property\undefined
\newtheorem{property}{Property}
\fi

\ifx\lemma\undefined
\newtheorem{lemma}{Lemma}[section]
\fi

\ifx\proposition\undefined
\newtheorem{proposition}{Proposition}[section]
\fi

\ifx\remark\undefined
\newtheorem{remark}{Remark}
\fi

\ifx\corollary\undefined
\newtheorem{corollary}{Corollary}[section]
\fi

\ifx\definition\undefined
\newtheorem{definition}{Definition}[section]
\fi

\ifx\conjecture\undefined
\newtheorem{conjecture}{Conjecture}[section]
\fi

\ifx\axiom\undefined
\newtheorem{axiom}[theorem]{Axiom}
\fi

\ifx\claim\undefined
\newtheorem{claim}[theorem]{Claim}
\fi

\ifx\assumption\undefined
\newtheorem{assumption}{Assumption}
\fi

\ifx\problem\undefined
\newtheorem{problem}{Problem}[section]
\fi

\ifx\fact\undefined
\newtheorem{fact}{Fact}
\fi

\newcommand{\ie}{\textit{i.e.}\ }

\newcommand{\methodtitle}{EfficientRollout}

\newcommand{\this}[1]{\textcolor{blue}{what?}}

\newcommand{\nosd}{veRL (AR)}
\newcommand{\alwayssd}{quantized self-SD}
\newcommand{\Alwayssd}{Quantized self-SD}

\newcommand{\history}{history-based drafting}
\newcommand{\History}{History-based drafting}
\newcommand{\learned}{learned auxiliary drafting}
\newcommand{\Learned}{Learned auxiliary drafting}

\newcommand{\TargetInduced}{Target-induced drafting}
\newcommand{\nonparametric}{non-parametric drafting}

\newcommand{\speedupete}{15.6\%}

\newcommand{\speeduprollout}{24.2\%}

\newcommand{\DrafterDist}{q}
\newcommand{\TargetDist}{p}
\newcommand{\Drafter}{\mathcal{M}_{\DrafterDist}}
\newcommand{\Target}{\mathcal{M}_{\TargetDist}}
\newcommand{\TargetQuantized}{Q(\Target)}
\newcommand{\DraftLength}{\gamma}
\newcommand{\TargetTime}{T_\TargetDist}
\newcommand{\TargetTimebs}{T_\TargetDist(B,S)}
\newcommand{\DraftTime}{T_\DrafterDist}
\newcommand{\DraftTimebs}{T_\DrafterDist(B,S)}
\newcommand{\VerifyTime}{T_V}
\newcommand{\VerifyTimebsg}{T_V(B,S,\DraftLength)}
\newcommand{\SDBlockTime}{T_{\mathrm{SD}}}
\newcommand{\SDSpeedup}{\mathrm{Speedup}_{\mathrm{SD}}}
\newcommand{\mal}{\tau}
\newcommand{\MAL}{block efficiency}
\newcommand{\bottleneck}{latency bottleneck}

\newcommand{\Tdecode}{T_{\mathrm{decode}}}
\newcommand{\Tdense}{T_{\mathrm{dense}}}
\newcommand{\Tattn}{T_{\mathrm{attn}}}

\newcommand{\CompFLOPs}{C_{\mathrm{FLOPs}}}
\newcommand{\MemBytes}{M_{\mathrm{Bytes}}}

\newcommand{\EffBandwidth}{\mathrm{BW}_{\mathrm{eff}}}
\newcommand{\RooflineTime}{T_{\mathrm{roofline}}}

\newcommand{\togglepolicy}{SD toggle policy}

\newcommand{\ToggleMargin}{\epsilon}
\newcommand{\TargetMemory}{M_T}
\newcommand{\DrafterMemory}{M_D}
\newcommand{\ForwardCompute}{C}
\newcommand{\TargetWeights}{W_T}
\newcommand{\DrafterWeights}{W_D}
\newcommand{\KVTrafficCoeff}{\kappa_{\mathrm{eff}}}
\newcommand{\DrafterQuantCoeff}{\eta_D}
\newcommand{\TargetOverhead}{c_T}
\newcommand{\DrafterOverhead}{c_D}
\newcommand{\VerifyOverhead}{c_V}
\newcommand{\DenseCompute}{C_{\mathrm{dense}}}
\newcommand{\AttnCompute}{C_{\mathrm{attn}}}
\newcommand{\SDToggle}{\pi_{\mathrm{SD}}}
\newcommand{\SDOn}{\texttt{sd\_on}}

\newcommand{\adaptivepolicy}{adaptive draft-length policy}

    \PassOptionsToPackage{numbers, sort&compress}{natbib}

 \usepackage[main, final]{neurips_2026}

\usepackage[utf8]{inputenc} 
\usepackage[T1]{fontenc}    
\usepackage[hidelinks]{hyperref}       
\usepackage{url}            
\usepackage{booktabs}       
\usepackage{amsfonts}       
\usepackage{nicefrac}       
\usepackage{microtype}      
\usepackage[table]{xcolor}         
\usepackage{makecell}
\usepackage{graphicx}
\usepackage{amsmath}
\usepackage{algorithm}
\usepackage[noend]{algpseudocode}
\usepackage{subcaption}
\usepackage{multirow}
\usepackage{caption}
\usepackage{comment}
\usepackage[capitalise]{cleveref}
\usepackage{enumitem}
\usepackage{titletoc}
\usepackage{tablefootnote}
\usepackage{threeparttable}
\usepackage{wrapfig}
\usepackage{mfirstuc}
\usepackage{duckuments}
\usepackage{wrapfig}
\usepackage{placeins}

\crefname{section}{Sec.}{Secs.}
\crefname{subsection}{Sec.}{Secs.}
\crefname{subsubsection}{Sec.}{Secs.}
\crefname{table}{Tab.}{Tabs.}

\Crefname{section}{Section}{Sections}
\Crefname{subsection}{Section}{Sections}
\Crefname{subsubsection}{Section}{Sections}
\Crefname{table}{Table}{Tables}

\title{\methodtitle{}: System-Aware Self-Speculative Decoding for RL Rollouts}

\author{%
\textbf{Minseo Kim}$^{1,2}$\thanks{Equal contribution.}\kern0.35em\thanks{Work done during an internship at FuriosaAI.}  \quad
\textbf{Minjae Lee}$^{1}$\footnotemark[1]  \quad
\textbf{Seunghyuk Oh}$^{1}$ \enspace
\textbf{Kevin Galim}$^{1}$ \enspace
\textbf{Donghoon Kim}$^{1}$ \\
\textbf{Coleman Hooper}$^{2}$ \enspace
\textbf{Harman Singh}$^{2}$ \enspace
\textbf{Amir Gholami}$^{2}$ \enspace
\textbf{Hyung Il Koo}$^{1}$ \enspace
\textbf{Wonjun Kang}$^{1}$ \\[0.4em]
$^{1}$FuriosaAI \quad $^{2}$University of California, Berkeley \\[0.2em]
\texttt{minseo.kim@berkeley.edu} \quad
\texttt{\{minjae.lee,kangwj1995\}@furiosa.ai}
}

\begin{document}

\maketitle
\setcounter{footnote}{0}

\begin{abstract}
Reinforcement learning (RL) has become a representative post-training paradigm for large language models (LLMs), enabling strong reasoning and agentic capabilities.
However, rollout generation remains a dominant \bottleneck{} because autoregressive (AR) sampling decodes responses sequentially and a small number of long-tailed generations often determine completion time.
Speculative decoding (SD) offers a natural way to address this bottleneck, as it is a well-established technique for serving fixed LLMs that reduces latency by rapidly drafting tokens and accepting them through parallel verification while preserving the target-model distribution.
However, its practical speedups do not directly carry over to RL rollouts: (i) the evolving target policy makes any fixed drafter increasingly mismatched with the policy's output distribution; and (ii) active batch sizes shrink throughout rollout decoding, shifting decoding from compute-bound to memory-bound regimes where parallel verification can exploit underutilized compute.
Therefore, accelerating RL rollouts requires both a drafter that remains effective under long, high-temperature generations from an evolving policy and system-aware use of SD that avoids compute-bound regimes.
We present \textbf{\methodtitle{}}, a system-aware self-SD framework designed to address this gap for RL rollouts.
\methodtitle{} induces a quantized drafter from the target model (\ie{}self-speculative decoding), keeping it coupled to the evolving policy without separate drafter pretraining or online adaptation.
It further coordinates a system-aware SD toggle policy with acceptance-aware draft-length adaptation, enabling speculation only in beneficial regimes while matching the drafting budget to evolving drafter quality.
\methodtitle{} reduces rollout and end-to-end latency by up to~\speeduprollout{} and~\speedupete{}, respectively, over an accelerated AR rollout baseline, while preserving final model quality.
\footnote{Code is available at \url{https://github.com/furiosa-ai/EfficientRollout}.}
\end{abstract}

\section{Introduction}
\label{sec:introduction}

Reinforcement learning (RL) has become an essential method for post-training frontier large language models (LLMs), enhancing their reasoning, math, and agentic capabilities~\citep{guo2025deepseek, yang2025qwen3, team2025kimi, zeng2025glm}.
Among RL algorithms, on-policy RL collects rollouts from the same policy being optimized, reducing policy-staleness degradation and improving training stability and model quality~\citep{kumar2025llm,fu2025areal}.
However, rollout generation has emerged as the dominant \bottleneck{} in the RL pipeline because post-trained models increasingly generate long responses~\citep{chen2024not,guo2025deepseek,yu2025dapo,liu2025understanding} that must be decoded token by token.
Consequently, reducing rollout latency is critical for accelerating RL post-training, since it often dominates other parallelizable stages such as policy updates~\citep{schulman2017proximal}.

Speculative decoding (SD)~\citep{leviathan2023fast,chen2023accelerating} offers a promising way to reduce inference latency while preserving the policy distribution.
SD uses a cheaper drafter to propose multiple candidate tokens, which are then verified in parallel by the target model, \textit{i.e.}, the policy being sampled.
Its speedup depends on two conditions: algorithmically, the block efficiency $\mal$ must be sufficiently high, where $\mal$ measures how many target-distributed tokens each draft-verify iteration produces on average~\citep{zhou2024distillspec,liu2025speculative};
system-wise, decoding should not be pushed into a compute-bound regime, so that parallel target-model verification can exploit underutilized compute~\citep{leviathan2023fast,sadhukhan2024magicdec,liu2025speculative}.

\begin{figure*}[t]
    \centering
    \includegraphics[width=\textwidth]{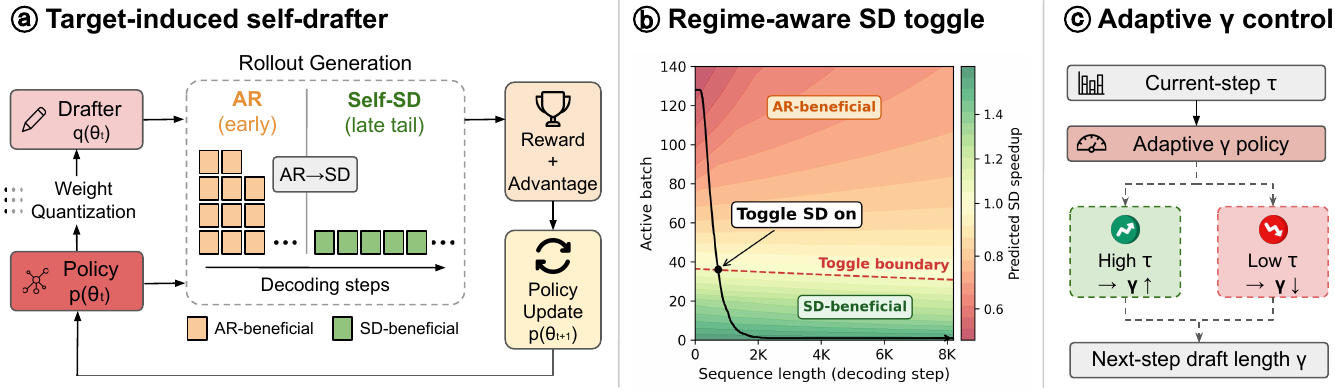}
    \caption{
    \textbf{Overview of \methodtitle{}}.
    \methodtitle{} bridges the gap between SD for fixed-model serving and RL rollout decoding through three coordinated components:
    (a) per-step self-drafter refresh to track the evolving policy,
    (b) regime-aware toggling from AR to SD under tail-heavy rollout dynamics (see~\cref{fig:rollout_bottleneck_appendix}), and
    (c) adaptive draft length $\gamma$ control based on observed block efficiency $\tau$.
    }
    \label{fig:main_teaser}
\end{figure*}

In this regard, RL introduces challenges absent from serving fixed LLMs, making existing SD methods nontrivial to apply to RL rollouts.
First, the target policy continuously evolves during training, making static drafters stale and potentially reducing block efficiency~\citep{zhang2025fastgrpo}.
Second, rollout generation often starts with large active batch sizes, creating compute-bound intervals where SD may be slower than autoregressive (AR) decoding, before the batch later shrinks as shorter responses finish.
At the same time, RL creates an opportunity: as post-training sharpens the policy distribution~\citep{zhao2025echo}, the target policy can be approximated by a lower-capacity subset induced from the target itself, which can still cover the concentrated probability mass.
Together, these factors raise a practical question: \emph{how can SD realize speedup in RL rollouts under the distinct conditions introduced by RL?}

Prior SD methods for RL rollouts address this problem, but still face either low block efficiency or the burden of drafter construction and adaptation.
\History{} methods reuse previous-epoch rollouts as draft proposals and are easy to deploy, but their low-capacity drafters often yield limited block efficiency because past rollouts sparsely cover future trajectories~\citep{liu2025spec,he2025history,shao2025beat}.
\Learned{} methods seek higher block efficiency by training auxiliary drafters, but existing learned auxiliary drafters do not automatically align with RL rollout distributions; in practice, they often require separate drafter pretraining and continual online adaptation to track the evolving target policy, adding training overhead and system complexity~\citep{zhang2025fastgrpo,chen2025respec,hu2026taming}.

To this end, we answer this question with \textbf{\methodtitle}, a system-aware SD framework designed for on-policy RL rollouts~(\cref{fig:main_teaser}).
We first characterize rollout workloads, showing that target-induced drafters naturally stay coupled to the evolving policy and that dense weight projections dominate tail decoding.
Based on these observations, we induce a weight-quantized drafter from the current target policy at every training step, achieving high block efficiency without separate drafter pretraining or online drafter updates.
We further introduce two control policies: a system-aware SD toggle that activates SD only in favorable regimes, avoiding slowdowns during early compute-bound large-batch phases, and an adaptive draft-length policy that matches the drafting budget to evolving acceptance behavior during RL.
We show that \methodtitle{} reduces rollout and end-to-end latency by up to~\speeduprollout{} and~\speedupete{}, respectively, over prevalent RL training and serving stacks~\citep{sheng2025hybridflow,kwon2023efficient}.

Our main contributions are as follows:

\begin{itemize}[leftmargin=*]
\item We characterize SD for RL rollouts, showing how evolving policies, shrinking-batch dynamics, and rollout-tail latency motivate a system-aware self-SD design~(\cref{sec:preliminary}).
\item We present \methodtitle{}, an RL-specific SD framework with a target-induced quantized drafter, dynamic SD toggling, and adaptive drafting budgets~(\cref{sec:method}).
\item We show that \methodtitle{} reduces rollout and end-to-end latency by up to~\speeduprollout{} and~\speedupete{} over a standard accelerated AR rollout baseline~\citep{sheng2025hybridflow,kwon2023efficient}, and analyze why existing SD approaches do not consistently resolve the emergent challenges of RL rollouts~(\cref{sec:Evaluation}).
\end{itemize}

\section{Related Work}
\label{sec:related-work}

\subsection{Reinforcement Learning for LLMs}

Frontier LLMs post-trained with RL have achieved state-of-the-art reasoning and agentic capabilities~\citep{guo2025deepseek,yang2025qwen3,team2025kimi,zeng2025glm}, enabled by advances in policy optimization~\citep{schulman2017proximal,shao2024deepseekmath,yu2025dapo,chen2025minimax}.
In particular, RL with verifiable rewards (RLVR) has become a dominant paradigm for reasoning-heavy domains such as math and coding, where supervision can be obtained from rule-based checkers or unit tests rather than learned reward models~\citep{wen2025reinforcement,guo2025deepseek}.
In on-policy RL, the policy being optimized is the same policy used for rollout, which avoids degradation from policy staleness and improves training stability and model quality~\citep{kumar2025llm,fu2025areal}.
However, rollout generation is often the main \bottleneck{} because RL tends to induce longer outputs~\citep{chen2024not,liu2025understanding}, amplifying the cost of sequential, memory-bandwidth-bound AR decoding that can leave compute underutilized~\citep{gholami2024ai}.
By contrast, other stages of the RL loop, including log-probability computation and policy updates, are more parallelizable.

\subsection{Speculative Decoding for LLM Inference}

SD accelerates target LLM decoding through a sequential drafting and parallel verification scheme~\citep{leviathan2023fast,chen2023accelerating}.
This draft-and-verify structure preserves the target-model distribution through rejection sampling while exploiting underutilized compute in memory-bound decoding regimes.
However, in compute-bound regimes, SD can be slower than standard AR decoding because parallel verification has little underutilized compute to exploit~\citep{sadhukhan2024magicdec}.
SD methods differ mainly in how they construct the drafter~\citep{xia2024unlocking}.
Canonical SD uses an independent smaller drafter model to approximate the target model~\citep{leviathan2023fast,chen2023accelerating}.
To make drafting much cheaper, \nonparametric{} methods generate drafts from text using n-gram, prefix-tree, or suffix-tree matching~\citep{he2024rest,ou2024lossless,stewart2024n}.
\Learned{} methods attach lightweight modules to the target model for efficient token prediction, such as FFN heads or one-layer drafters~\citep{li2024eagle,cai2024medusa}.
Self-speculative decoding (self-SD) instead derives the drafter directly from the target model via layer skipping~\citep{zhang2024draft}, sparse attention~\citep{yue2026specattn}, or quantization~\citep{tiwari2025quantspec}.

\subsection{Speculative Decoding for Rollout in Reinforcement Learning}

\begin{table}[t]
\centering
\small
\setlength{\tabcolsep}{6pt}
\renewcommand{\arraystretch}{1.18}
\caption{
Comparison of drafter categories for SD in RL rollouts.
We compare methods along three dimensions: whether they use parameterized drafters, avoid continuous online adaptation, and can accelerate from the first epoch without category-specific warm-up, such as rollout-history collection or drafter pretraining.
}
\label{tab:rl_sd_taxonomy}
\vspace{1mm}
\resizebox{1\linewidth}{!}{%
\begin{tabular}{
p{3.2cm}
!{\color{gray!45}\vrule width 0.4pt}
c
c
c
!{\color{gray!45}\vrule width 0.4pt}
>{\raggedright\arraybackslash}p{3.2cm}
}
\toprule
\textbf{Drafter category}
&
\textbf{Parameterized}
&
\textbf{Online-adaptation free}
&
\textbf{Warm-up free}
&
\textbf{Examples} \\
\midrule

\textbf{(1) History-based}
&
No
&
Yes
&
No
&
{\footnotesize \citep{liu2025spec,shao2025beat,he2025history}} \\[2pt]

\textbf{(2) Learned auxiliary}
&
Yes
&
No
&
Conditional
&
{\footnotesize \citep{zhang2025fastgrpo,hu2026taming,chen2025respec,iso2026accelerating}} \\[2pt]

\rowcolor{gray!10}
\textbf{(3) Target-induced}
&
\textbf{Yes}
&
\textbf{Yes}
&
\textbf{Yes}
&
{\footnotesize \bfseries \methodtitle{} (ours)} \\

\bottomrule
\end{tabular}%
}
\end{table}

Several prior works use SD to accelerate rollout, a major RL \bottleneck{}.
\Cref{tab:rl_sd_taxonomy} summarizes the main drafting methods for RL-specific SD:
\textbf{(1) \History{}} methods adapt \nonparametric{} by reusing rollouts from previous epochs as draft proposals, e.g., via prefix- or suffix-based matching~\citep{liu2025spec,he2025history,shao2025beat}.
They are easy to deploy and require no parameterized drafter, but often yield short accepted drafts because past trajectories sparsely cover future rollouts. They also require a warm-up period to collect rollout history and may rely on acceptance-boosting heuristics such as reduced temperatures or lossy lenience settings.
\textbf{(2) \Learned{}} methods adapt auxiliary-drafter approaches from fixed-model SD, often using EAGLE-style drafters~\citep{li2024eagle,li2025eagle}, and aim to match the drafter to long-reasoning rollout distributions~\citep{zhang2025fastgrpo,chen2025respec,hu2026taming,iso2026accelerating}.
However, such drafters are not generally available as public checkpoints, so obtaining them often requires dedicated drafter pretraining or several warm-up epochs before achieving effective speedup~\citep{zhang2025fastgrpo}.
Tracking the evolving RL policy also requires online adaptation, adding system complexity and management overhead~\citep{hu2026taming}.
\textbf{(3) \TargetInduced{}} methods correspond to self-SD methods that derive the drafter solely from the current target model.
Our work instantiates this category with \textbf{\methodtitle{}}, which uses a weight-quantized copy of the target model as the drafter, staying synchronized with the evolving policy without separate online adaptation while maintaining high block efficiency.
We focus on lossless on-policy RL acceleration and therefore exclude rollout-efficiency techniques that change the rollout distribution or termination process, such as lower-precision rollout inference~\citep{li2026qurl}, proxy rollout models~\citep{chen2026jackpot}, and early rollout termination~\citep{zhang2026sortedrl,zhou2025april}.

\section{Preliminary}
\label{sec:preliminary}

\subsection{Speedup Factors in Speculative Decoding}
\label{subsec:sd}

For a given batch size $B$ and sequence length $S$, let $\DraftTimebs$ and $\TargetTimebs$ be the single-token decoding times of the drafter $\Drafter$ and target model $\Target$, respectively.
When the drafter sequentially speculates $\DraftLength$ tokens, let $\VerifyTimebsg$ denote the parallel verification time by the target model.\footnote{For brevity, we omit $B$, $S$, and $\gamma$ when clear.}
We define block efficiency $\mal$ as the average number of accepted draft tokens plus one target-sampled token per drafting block~\citep{zhou2024distillspec}.
Under the i.i.d. acceptance assumption, $\mal$ can be written in terms of a per-token acceptance rate $\alpha$ as
$\mal=(1-\alpha^{\DraftLength+1})/(1-\alpha)$~\citep{leviathan2023fast,sadhukhan2024magicdec}.
Given $\mal$, the SD block time and wall-clock speedup are
\begin{equation}
\SDBlockTime(B,S,\gamma)
=
\gamma \DraftTimebs + \VerifyTimebsg,
\quad
\SDSpeedup(B,S,\gamma)
=
\mal \frac{\TargetTimebs}
{\SDBlockTime(B,S,\gamma)}.
\label{eq:sd_speedup}
\end{equation}
Thus, SD acceleration depends on both high $\mal$ and the latency ratios
$\DraftTime/\TargetTime$ and $\VerifyTime/\TargetTime$ induced by $(B,S)$.
When decoding becomes compute-bound, often at large $B$, verification has little underutilized compute to exploit.
This increases $\VerifyTime/\TargetTime$ and can make SD slower than AR decoding, an effect often hidden in common $B=1$ SD benchmarks~\citep{sadhukhan2024magicdec,xia2024unlocking}.

\subsection{Roofline Modeling for LLM Decoding Latency}
\label{subsec:roofline-latency}
To identify the decoding regime on the actual hardware system, we use roofline modeling, a standard system-aware approach~\citep{williams2009roofline,yuan2024llm}.
Roofline modeling distinguishes between arithmetic work $\CompFLOPs$, executed with effective compute capacity $\mathrm{F}_{\mathrm{eff}}$, and data movement $\MemBytes$, served by effective memory bandwidth $\EffBandwidth$.
Assuming computation and memory access can ideally overlap, the workload latency is approximated as
\begin{equation}
\RooflineTime
\approx
\max\left(
\frac{\MemBytes}{\EffBandwidth},
\frac{\CompFLOPs}{\mathrm{F}_{\mathrm{eff}}}
\right).
\label{eq:roofline_basic}
\end{equation}

For LLM decoding, $\MemBytes$ includes weight and KV-cache traffic, while $\CompFLOPs$ includes dense projections such as QKVO, FFN, and LM-head computations, as well as attention operations~\citep{sadhukhan2024magicdec}.

\subsection{RL Rollout Latency Bottlenecks from Long-Tail Decoding}
\label{subsec:rollout_bottleneck}

\begin{figure*}[t]
    \centering

    \begin{subfigure}[t]{0.32\linewidth}
        \centering
        \includegraphics[width=\linewidth]{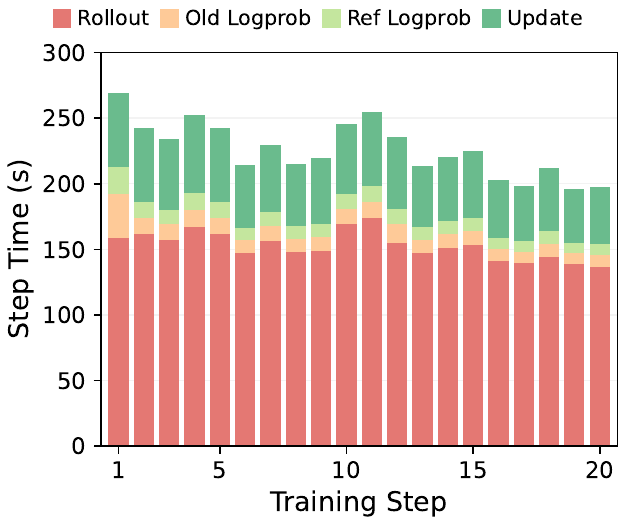}
        \caption{}
        \label{fig:rollout_llama_bottleneck_step}
    \end{subfigure}
    \hfill
    \begin{subfigure}[t]{0.32\linewidth}
        \centering
        \includegraphics[width=\linewidth]{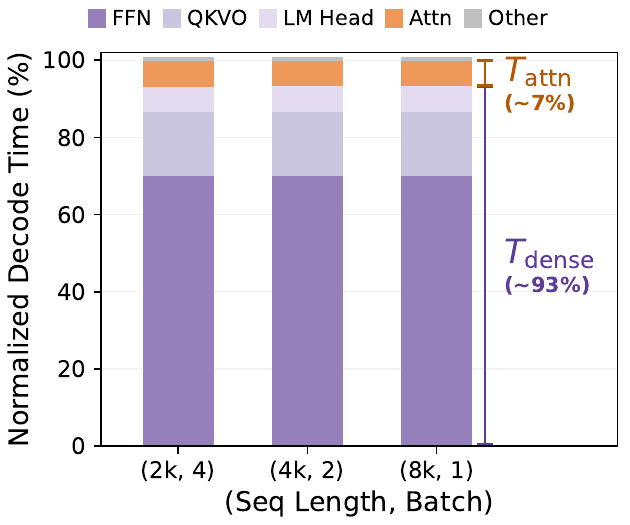}
        \caption{}
        \label{fig:rollout_llama_decode_breakdown}
    \end{subfigure}
    \hfill
    \begin{subfigure}[t]{0.32\linewidth}
        \centering
        \includegraphics[width=\linewidth]{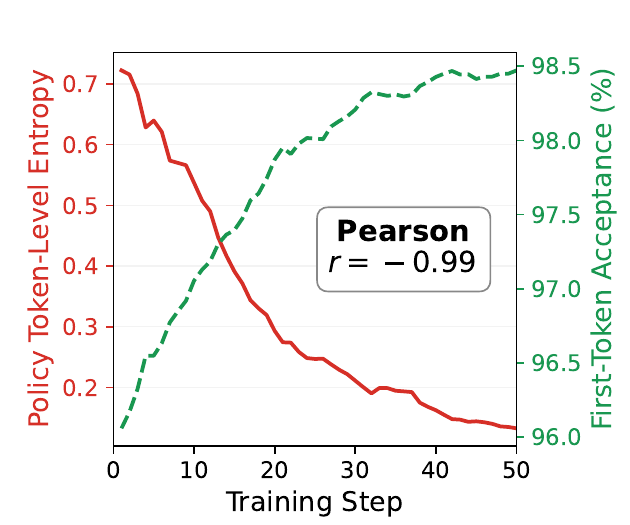}
        \caption{}
        \label{fig:entropy_acceptance_qwen}
    \end{subfigure}

    \caption{
    \textbf{Empirical characteristics of RL rollout decoding.}
    (a) Step-time decomposition over the first 20 training steps.
    (b) Single-token decode-time breakdown in RL rollout-tail phases.
    (c) Inverse correlation between target-policy entropy and quantized-drafter first-token acceptance over steps.
    }
    \label{fig:rollout_bottleneck}
\end{figure*}

Rollout is the primary \bottleneck{} in RL post-training for LLMs.
As shown in \cref{fig:rollout_llama_bottleneck_step}, sequential rollout decoding accounts for nearly 70\% of total latency on average.
This bottleneck is amplified in RL because post-training workloads increasingly produce long reasoning trajectories, and rollout generation must sample these trajectories token by token under AR decoding~\citep{gholami2024ai}.
By contrast, policy updates and log-probability computation are more parallelizable, allowing them to better utilize compute capacity.
Within a rollout, after shorter responses finish and the active batch shrinks, a small number of long responses often determine the makespan (\cref{fig:main_teaser}) while leaving compute underutilized.
This underutilized compute provides the runtime budget that SD can exploit.
\Cref{app:long_tail_analysis} provides additional analysis of the long-tail completion behavior of rollout during RL.

\section{{\methodtitle{}: From RL Rollout Dynamics to System-Aware Self-SD}}
\label{sec:method}

This section develops \textbf{\methodtitle{}} from three RL rollout characteristics.
First, the evolving target policy and rollout-tail latency profile motivate a weight-quantized drafter for self-SD.
Second, shifting runtime regimes motivate an SD toggle policy that activates only when beneficial.
Third, changing acceptance behavior during RL motivates adaptive draft lengths.
We describe each component below; \Cref{alg:full-pipeline} summarizes the full \methodtitle{} pipeline for~\cref{subsec:target-induced-drafter,subsec:sd-toggle,subsec:adaptive-gamma}.

\subsection{Weight-Quantized Drafter for Self-Speculative Decoding}
\label{subsec:target-induced-drafter}

\textbf{Self-speculative decoding.}
Among the SD designs in \cref{tab:rl_sd_taxonomy}, self-SD with a target-induced drafter~\citep{zhang2024draft,yue2026specattn,tiwari2025quantspec} is well suited to RL, where the target model evolves throughout training.
Because the drafter is derived directly from the current target, it remains synchronized with the evolving policy without separate drafter pretraining or online adaptation.
This contrasts with learned auxiliary drafters, which can become stale unless continually adapted online to track the evolving policy~\citep{zhang2025fastgrpo}.

\textbf{Rollout-tail latency decomposition.}
We decompose rollout-tail single-token decoding latency, $\Tdecode$, to identify which component a self-SD drafter should make cheaper.
As shown in \cref{fig:rollout_llama_decode_breakdown}, dense projection time $\Tdense$, including QKVO, FFN, and LM-head projections, accounts for around 90\% of total latency, exceeding attention time $\Tattn$ by more than $10\times$.
This observation is consistent with modern LLM architectures that reduce KV-cache sizes~\citep{shazeer2019fast,ainslie2023gqa,liu2024deepseek,guo2025deepseek} and with system analyses showing that parameter loading dominates latency in small-batch memory-bound regimes~\citep{sadhukhan2024magicdec}.

\textbf{Weight-quantized drafting.}
We induce a weight-quantized drafter~\citep{tiwari2025quantspec} from the current target model to reduce the weight-loading cost of $\Tdense$, the dominant component of rollout-tail latency (\cref{fig:rollout_llama_decode_breakdown}).
This lowers the drafter-to-target latency ratio $\DraftTime/\TargetTime$ in~\cref{eq:sd_speedup}.
Specifically, we apply lightweight RTN quantization to the FFN and QKVO projection layers at the beginning of each training step, following prior practice~\citep{lin2024awq}.
We use 4-bit weights (W4) as a practical latency--acceptance trade-off: lower precision makes drafting substantially cheaper while preserving sufficient block efficiency $\mal$ for effective SD.
In contrast, sparse-attention drafting, another representative self-SD approach, mainly targets $\Tattn$, which contributes far less than $\Tdense$ to rollout-tail latency in this regime.
Layer skipping is also a self-SD option; however, fixed skipping patterns often provide limited practical speedup~\citep{xia2024unlocking,song2026knn}, while adaptive skipping is difficult to combine with static inference optimizations in modern serving engines, such as CUDA graph pre-capture~\citep{xia2024swift,chen2025clasp,kwon2023efficient}.
\Cref{app:w4_vs_w8,app:practical_quantized_vs_sparse} further analyze self-SD alternatives and W4--W8 latency--acceptance trade-off.

\subsection{Regime-Aware SD Toggle Policy via Roofline Modeling}
\label{subsec:sd-toggle}

Even with a cheap quantized drafter, SD is not beneficial throughout the entire rollout.
Early rollout phases have large active batches that can saturate compute, leaving little underutilized compute for target verification (\ie{}high $\VerifyTime/\TargetTime$ in~\cref{eq:sd_speedup}). As a result, SD can be slower than AR decoding despite high $\mal$.
As shorter responses finish, the active batch shrinks and decoding enters a low-batch tail where SD becomes beneficial.
To decide when to enable SD, we use a roofline model to predict speedup (\cref{subsec:roofline-latency}) from the current decoding state $(B,S)$ and SD runtime state $(\mal,\gamma)$, activating SD only when the prediction is favorable.

\textbf{SD toggle policy.} 
To implement this decision, the SD toggle policy uses a calibrated model to predict the SD-over-AR speedup for the current decoding state $(B,S)$.
We first define memory-side and compute-side cost estimates for target, draft, and verification execution:
\begin{equation*}
\TargetMemory
=
\frac{\TargetWeights+\KVTrafficCoeff BS}{\EffBandwidth},
\quad
\DrafterMemory
=
\frac{\DrafterQuantCoeff\DrafterWeights+\KVTrafficCoeff BS}{\EffBandwidth},
\quad
\ForwardCompute
=
\frac{B(\DenseCompute+S\AttnCompute)}{\mathrm{F}_{\mathrm{eff}}}.
\end{equation*}
Here, $\TargetMemory$ and $\DrafterMemory$ estimate data-movement costs for one target or quantized-drafter forward pass, including weight and effective KV-cache traffic, while $\ForwardCompute$ estimates the arithmetic cost of one forward pass.
$\TargetWeights$ and $\DrafterWeights$ denote the target and drafter weight sizes, respectively, and $\DenseCompute$ and $\AttnCompute$ denote architecture-determined dense and attention compute costs.
Because the target and self-drafter share the same architecture, they share the same compute term; the quantized drafter's reduced weight traffic is captured by $\DrafterQuantCoeff$ in the memory-side term.
Using~\cref{eq:roofline_basic}, we instantiate the roofline latency for target decoding, quantized-drafter decoding, and target verification as
\begin{equation*}
\widehat{\TargetTime}
=
\max\{\TargetMemory,\ForwardCompute\}
+
\TargetOverhead B,
\quad
\widehat{\DraftTime}
=
\max\{\DrafterMemory,\ForwardCompute\}
+
\DrafterOverhead B,
\quad
\widehat{\VerifyTime}
=
\max\{\TargetMemory,(\DraftLength+1)\ForwardCompute\}
+
\VerifyOverhead B.
\label{eq:toggle_latencies}
\end{equation*}
The overhead terms capture non-ideal overlap and batch-dependent residual costs beyond ideal roofline approximation.
Following the SD speedup model in~\cref{eq:sd_speedup}, we define the SD toggle policy $\SDToggle$ to activate SD when the predicted speedup exceeds a safety margin:
\begin{equation}
\SDToggle(B,S,\mal,\DraftLength)
=
\mathbf{1}
\left[
\frac{
\mal\,\widehat{\TargetTime}
}{
\DraftLength\widehat{\DraftTime}
+
\widehat{\VerifyTime}
}
\geq
1+\ToggleMargin
\right],
\qquad
\ToggleMargin\geq0.
\label{eq:sd_speedup_toggle}
\end{equation}
Once activated, SD remains enabled until the end of the rollout, as the active batch size monotonically decreases toward the SD-beneficial tail regimes.

\textbf{Calibrated parameters.}
We introduce additional calibration parameters to adapt the roofline model and the corresponding policy to different model and system configurations.
The fitted terms $\KVTrafficCoeff$, $\DrafterQuantCoeff$, and 
$\TargetOverhead,\DrafterOverhead,\VerifyOverhead$ capture effective KV-cache traffic, quantized-drafter effects, and residual per-batch overheads for target, draft, and verification execution, respectively.
These parameters are calibrated once per model--hardware pair using a pre-hoc profiling sweep and can be reused across RL runs. \Cref{app:toggle_calibration} provides calibration details.

\subsection{Adaptive Draft-Length Policy}
\label{subsec:adaptive-gamma}

\paragraph{Acceptance behavior evolves during RL.}
The weight-quantized drafter can become more effective over training, increasing $\tau$ as RL sharpens the target policy.
In \cref{fig:entropy_acceptance_qwen}, target-output entropy and token-level agreement between $\Target$ and its weight-quantized copy $\TargetQuantized$ show a strong negative correlation ($r=-0.99$).
As training progresses, RL can reduce target-output entropy~\citep{jin2025revisiting,cui2025entropy}, sharpening the target policy by concentrating probability mass on a smaller set of high-reward tokens~\citep{huang2024self,yang2025llm}.
A sharper distribution can make top-token decisions less sensitive to small quantization-induced perturbations, increasing agreement between $\Target$ and $\TargetQuantized$.
Consistent with this explanation, we observe that the quantized self-drafter's $\tau$ improves over training, motivating adaptive draft budgets that increase as acceptance improves.
\Cref{app:entropy_acceptance} provides full experimental results for policy sharpening and block-efficiency improvement.

\paragraph{Adaptive policy for draft length $\gamma$.}
To exploit the increasing $\tau$ observed during RL training, we choose the next draft length $\gamma_{t+1}$ using the measured block efficiency $\mal_t$ as a proxy for current drafter quality.
We maintain an ordered draft-length set $\Gamma=[\gamma_{\mathrm{low}},\dots,\gamma_{\mathrm{high}}]$ and update from $\gamma_t$ to $\gamma_{t+1}$ only when the condition on $\mal_t$ persists for $P$ consecutive training steps.
Specifically, we increase the draft length when $\mal_t$ is close to the current draft-length ceiling, and decrease it when acceptance is too low to amortize draft computation.
The patience window $P$ prevents transient fluctuations in $\mal_t$ across steps from changing the draft length.
This removes the need to select a single fixed $\DraftLength$ through post-hoc sweeps, instead letting the adaptive policy move within a pre-specified draft-length set using observed $\mal_t$ feedback during RL training.

\begin{algorithm}[t]
\caption{\methodtitle{} Pipeline}
\label{alg:full-pipeline}
{\textbf{Parameter}:} draft-length set $\Gamma=[\gamma_{\mathrm{low}},\dots,\gamma_{\mathrm{high}}]$, toggle margin $\ToggleMargin$, block-efficiency thresholds $\alpha_{\mathrm{up}},\alpha_{\mathrm{down}}$, patience $P$\\
{\textbf{Input}:} target model, rollout states with runtime batch size $B$, and sequence length $S$.
\begin{algorithmic}[1]
\State Initialize $\gamma_1 \gets \gamma_{\mathrm{low}}$ and $\mal_0 \gets \gamma_{\mathrm{low}}$
\For{training timestep $t=1,2,\dots$}
    \State Refresh the drafter by quantizing the current target model
    \State Initialize $\SDOn \gets \textbf{false}$
    \While{rollout not finished}
        \If{$\SDOn=\textbf{false}$ and $\SDToggle(B,S,\mal_{t-1},\gamma_t)=0$}
            \State Decode autoregressively
        \ElsIf{$\SDOn=\textbf{false}$ and $\SDToggle(B,S,\mal_{t-1},\gamma_t)=1$}
            \State $\SDOn \gets \textbf{true}$
        \Else
            \State Decode with SD using draft length $\gamma_t$
        \EndIf
    \EndWhile
    \State Measure step-level block efficiency $\mal_t$
    \State $\gamma_{t+1} \gets
    \begin{cases}
    \mathrm{next}_{\Gamma}(\gamma_t),
    & \min\limits_{t' \in [t-P+1,t]} \mal_{t'} \ge 1+\gamma_t\alpha_{\mathrm{up}},\ \gamma_t<\gamma_{\mathrm{high}}, \\
    \mathrm{prev}_{\Gamma}(\gamma_t),
    & \max\limits_{t' \in [t-P+1,t]} \mal_{t'} \le 1+\gamma_t\alpha_{\mathrm{down}},\ \gamma_t>\gamma_{\mathrm{low}}, \\
    \gamma_t,
    & \text{otherwise.}
    \end{cases}$
    \State Run standard post-rollout optimization
\EndFor
\end{algorithmic}
\end{algorithm}

\section{Experiments}
\label{sec:Evaluation}

\subsection{Setup}
\label{sec:eval-setting}

\textbf{Training details.}
We evaluate representative open-source models across model families and scales: Qwen2.5-\{7B,14B\}~\citep{Yang2024Qwen25TR} on SimpleRL-8k-hard, and Llama3.1-8B~\citep{grattafiori2024llama} on SimpleRL-8k-medium, following the math-domain RLVR recipe~\citep{zeng2025simplerl}.
We evaluate acceleration of RL with GRPO~\citep{shao2024deepseekmath}, using train batch size 128, group size 8, rollout temperature 1.0, and maximum response length 8k for 100 training steps.
Unless otherwise noted, experiments are conducted on a single node with 8 A100-80GB GPUs using data parallelism (DP)~\citep{shao2025beat}. Additional evaluations on H100, Qwen2.5-32B with TP$=2$ on A100, and broader workload settings are provided in~\cref{app:extended_e2e_evaluation}.

\textbf{System configuration.}
To ensure both deployability and measured speedup, we implement the weight-quantized drafter on top of veRL~\citep{sheng2025hybridflow} and vLLM~\citep{kwon2023efficient}.
Our integration uses the Marlin weight-only quantization kernel~\citep{frantar2025marlin}, which reduces weight-loading cost while preserving compatibility with production-level inference optimizations.
We use a fixed safety margin $\epsilon=0.05$ for \togglepolicy{} and a shared draft-length set $\Gamma=\{5,7,9,11\}$ for \adaptivepolicy{} across all models. Full details are provided in~\cref{app:training_method_config}.

\textbf{Baselines.}
We compare \methodtitle{} against four baselines.
First, \textbf{\nosd{}} denotes accelerated AR inference using the standard rollout backend built on veRL and vLLM~\citep{sheng2025hybridflow,kwon2023efficient}, without SD.
Second, the \textbf{\history{}} baseline evaluates Spec-RL~\citep{liu2025spec}, a rollout-history-based SD method using prefix matching, in the same rollout stack.
We use its best-performing lenience factor $e^{0.5}$, which is needed for meaningful speedup but makes the method lossy and deviate from the target distribution.
Third, the \textbf{\learned{}} baseline evaluates an EAGLE3~\citep{li2025eagle}-based auxiliary drafter implemented on top of veRL and vLLM, following the RL-SD pipelines of FastRL~\citep{hu2026taming} and NeMo RL~\citep{iso2026accelerating}.
We use a fixed draft length of $\DraftLength=3$, following the best-performing setting in~\citet{iso2026accelerating}.
Lastly, \textbf{\alwayssd{}} applies SD with the weight-quantized drafter from~\cref{sec:method}, with SD always enabled.
It uses the same production-level kernel optimization as \methodtitle{}, and we select the best fixed draft length $\DraftLength^{*} \in \{5,7,9,11\}$ for each configuration.
Additional details on the \history{} and \learned{} baselines are provided in~\cref{app:history_drafting,app:learned_auxiliary_drafting}.

\textbf{Metrics.}
We report per-step averaged latency metrics that separate SD-specific overhead, rollout latency, and end-to-end training time.
Preparation time (Prep.) measures the extra work required by each SD method, including rollout-history lookup, online drafter forward/backward passes and parameter updates, and per-step drafter quantization.
Rollout Gen. measures rollout-generation time, while Step Time measures end-to-end training-step time, including non-rollout components such as log-probability computation and actor updates.
$\mal$, $\bar{\gamma}$, and $\alpha$ denote the average block efficiency, average draft length, and per-token acceptance rate used to compare methods with different $\bar{\gamma}$ values, respectively.
Additional measurement details are provided in~\cref{app:metric_measurement}.

\begin{table*}[t]
\centering
\footnotesize
\setlength{\tabcolsep}{3.6pt}
\renewcommand{\arraystretch}{1.08}
\caption{
End-to-end training acceleration across models and rollout decoding methods.
We report SD-specific preparation time, average block efficiency $\mal$ ($\bar{\gamma}$ denotes the average draft length), per-token acceptance rate $\alpha$, rollout-generation time, and training-step time.
Signed percentages denote relative changes from the \nosd{} baseline.
In arrow-marked columns, bold and underlined entries indicate the \textbf{best} and \underline{second-best} values within each model group.
}
\label{tab:e2e_speedup}
\begin{tabular}{llccccc}
\toprule
\textbf{Model}
&
\textbf{Drafting Method}
&
\textbf{Prep. (s) $\downarrow$}
&
\textbf{$\mal$ / $\bar{\gamma}$}
&
\textbf{$\alpha$ (\%) $\uparrow$}
&
\textbf{Rollout Gen. (s) $\downarrow$}
&
\textbf{Step Time (s) $\downarrow$} \\
\midrule

\multirow{5}{*}{Qwen2.5-7B}
& \nosd{}
& --
& --
& --
& 82.4
& 132.6 \\
& History-based$^\dagger$
& \textbf{1.1}
& N/A
& N/A
& 86.5 (+4.9\%)
& 133.8 (+0.9\%) \\
& Learned auxiliary
& 2.2
& 2.1 / 3.0
& 57.3
& 81.9 (-0.6\%)
& 132.1 (-0.4\%) \\
& \Alwayssd{}
& \underline{1.3}
& 7.5 / 7.0
& \textbf{98.1}
& \underline{66.1 (-19.7\%)}
& \underline{115.7 (-12.8\%)} \\
\rowcolor{gray!10}
\cellcolor{white}
& \methodtitle{}
& \underline{1.3}
& 8.6 / 8.3
& \textbf{98.1}
& \textbf{62.4 (-24.2\%)}
& \textbf{111.9 (-15.6\%)} \\

\midrule

\multirow{5}{*}{Qwen2.5-14B}
& \nosd{}
& --
& --
& --
& 126.6
& 221.1 \\
& History-based$^\dagger$
& \textbf{1.1}
& N/A
& N/A
& 129.5 (+2.3\%)
& 218.9 (-1.0\%) \\
& Learned auxiliary
& \underline{2.1}
& 2.2 / 3.0
& 60.3
& \underline{115.4 (-8.9\%)}
& \underline{213.8 (-3.3\%)} \\
& \Alwayssd{}
& 2.6
& 5.6 / 5.0
& \underline{97.3}
& 131.4 (+3.8\%)
& 223.2 (+0.9\%) \\
\rowcolor{gray!10}
\cellcolor{white}
& \methodtitle{}
& 2.6
& 7.4 / 7.0
& \textbf{97.6}
& \textbf{104.9 (-17.1\%)}
& \textbf{197.7 (-10.6\%)} \\

\midrule

\multirow{5}{*}{Llama3.1-8B}
& \nosd{}
& --
& --
& --
& 126.4
& 186.9 \\
& History-based$^\dagger$
& \textbf{1.1}
& N/A
& N/A
& 131.1 (+3.7\%)
& 187.4 (+0.2\%) \\
& Learned auxiliary$^\ddagger$
& 2.6
& 2.0 / 3.0
& 54.3
& 172.8 (+36.7\%)
& 234.8 (+25.6\%) \\
& \Alwayssd{}
& \underline{1.4}
& 5.5 / 5.0
& \textbf{96.4}
& \underline{116.7 (-7.6\%)}
& \underline{178.1 (-4.7\%)} \\
\rowcolor{gray!10}
\cellcolor{white}
& \methodtitle{}
& \underline{1.4}
& 5.3 / 5.0
& \underline{95.1}
& \textbf{112.2 (-11.2\%)}
& \textbf{173.3 (-7.3\%)} \\

\bottomrule
\end{tabular}

\vspace{2pt}
\begin{minipage}{0.98\linewidth}
\footnotesize
$^\dagger$ For \history{}, $\mal$ and $\alpha$ are not directly comparable; prefix-reuse statistics are reported in~\cref{app:why-history-slow}.

$^\ddagger$ We further analyze the Llama3.1-8B slowdown of \learned{} in~\cref{app:why-learned-auxiliary-challenging}.
\end{minipage}
\end{table*}

\begin{figure*}[t]
    \centering

    \begin{minipage}[t]{0.54\textwidth}
        \centering
        \includegraphics[width=\linewidth]{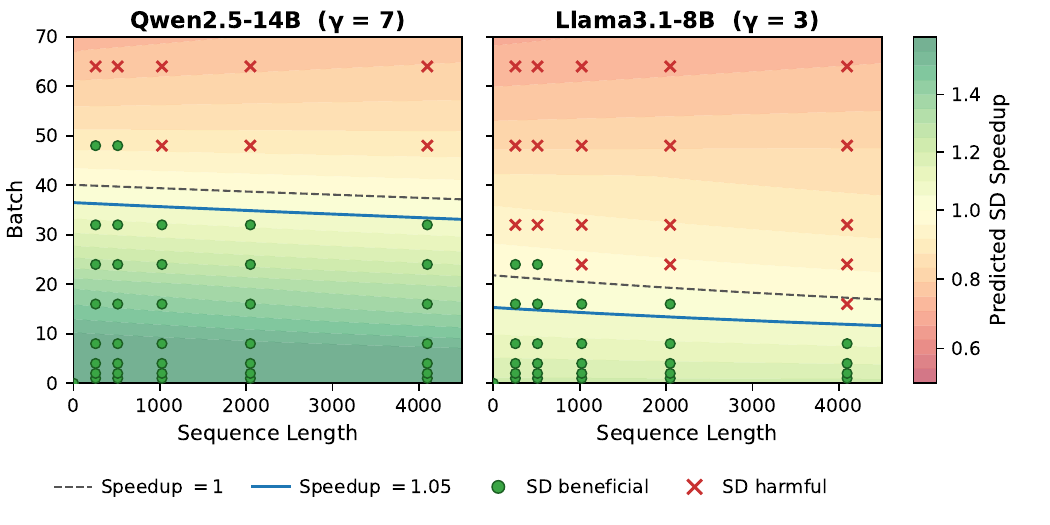}
        \caption{
        \textbf{Validation of the roofline-based toggle boundary.}
        Colors show the predicted speedup over the batch-size and sequence-length plane, and markers indicate whether measured SD is beneficial or harmful at the corresponding coordinates.
        }
        \label{fig:toggle_boundary}
    \end{minipage}
    \hfill
    \begin{minipage}[t]{0.42\textwidth}
        \centering
        \includegraphics[width=\linewidth]{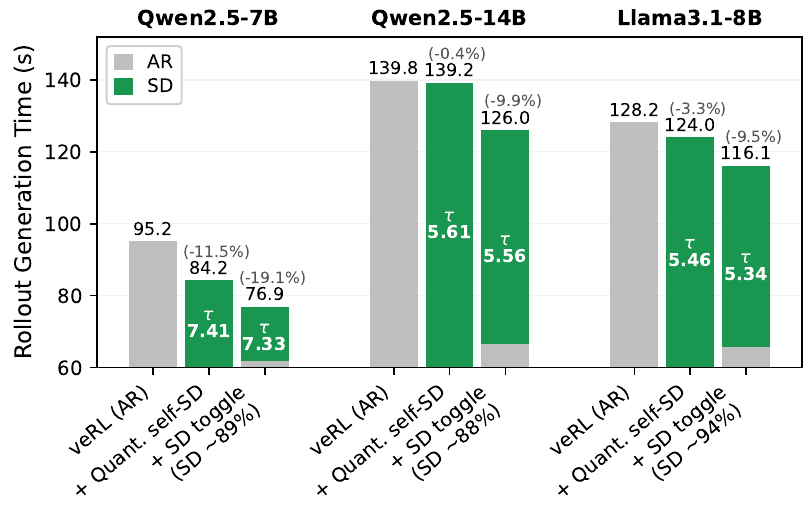}
        \caption{
        \textbf{Effect of regime-aware SD toggling.}
        Average rollout time over the first 50 training steps, showing that toggling avoids harmful early-regime speculation across all evaluated models.
        }
        \label{fig:toggle_vs_alwayson}
    \end{minipage}

    \vspace{-0.5em}
\end{figure*}

\subsection{End-to-End RL Training Acceleration}
\label{subsec:eval-e2e-acceleration}

\textbf{End-to-end latency.}
\Cref{tab:e2e_speedup} shows that \methodtitle{} achieves the largest latency reduction among all methods for every evaluated model.
The per-step quantization overhead is modest (1.3--2.6\,s) relative to the rollout-time savings: \methodtitle{} reduces rollout-generation latency by up to \speeduprollout{}, yielding an end-to-end training-step latency reduction of up to \speedupete{} after accounting for all RL operations.
We further quantify the individual contributions of W4 self-drafting, the Roofline toggle, and adaptive $\gamma$ to rollout-generation and full-step latency reductions in~\cref{app:component_decomposition}.
The \history{} baseline has the lowest preparation overhead from history lookup (1.1\,s), but does not reduce rollout-generation latency.
Even after rollout history becomes available, it reuses only 4.4--50.2\% of response tokens while adding 10.7--19.5\,s of verification overhead (\cref{tab:history_based_slow}).
Intuitively, prefix reuse acts like a large draft-and-verify block over the full candidate sequence, and even the highest reuse rate (50.2\%) is insufficient to amortize the added verification cost in our setting.
The \learned{} baseline shows model-dependent, limited gains because its auxiliary drafter does not consistently provide sufficient block efficiency to offset SD overheads (\cref{subsec:further-analysis}).
For example, it reduces Qwen2.5-14B step latency by 3.3\% but increases Llama3.1-8B step latency by 25.6\%.
Finally, \alwayssd{} reduces step latency on Qwen2.5-7B and Llama3.1-8B, but the Qwen2.5-14B slowdown shows that the quantized self-drafter alone is not sufficient for robust speedups.
We provide detailed analyses of slowdowns in the \history{} and \learned{} baselines in~\cref{app:why-history-slow,app:why-learned-auxiliary-challenging}.

\textbf{Acceptance rate and block efficiency.}
\Cref{tab:e2e_speedup} shows that \methodtitle{} attains consistently high $\alpha$ (95.1--98.1\%) together with large $\mal$.
\Alwayssd{} also maintains high $\alpha$ across models, indicating strong alignment between the target model and the quantized self-drafter.
\methodtitle{} preserves this high $\alpha$ while using larger $\bar{\gamma}$, indicating that adaptive $\gamma$ control can exploit training-time improvements in acceptance as it increases the draft budget.
By contrast, the \learned{} baseline remains limited to much lower $\alpha$ (54.3--60.3\%) even with shorter $\bar{\gamma}$, reflecting limited auxiliary-drafter effectiveness in long, high-temperature RL rollouts (\cref{subsec:further-analysis}).

\begin{figure}[t]
    \centering

    \begin{subfigure}[t]{0.32\linewidth}
        \centering
        \includegraphics[width=\linewidth]{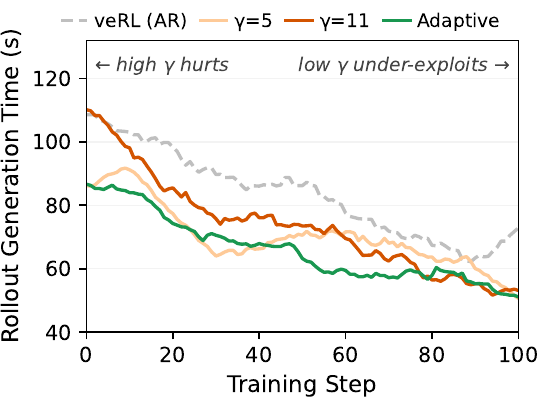}
        \caption{Rollout generation time}
        \label{fig:adaptive_gamma_gen_time}
    \end{subfigure}
    \hfill
    \begin{subfigure}[t]{0.32\linewidth}
        \centering
        \includegraphics[width=\linewidth]{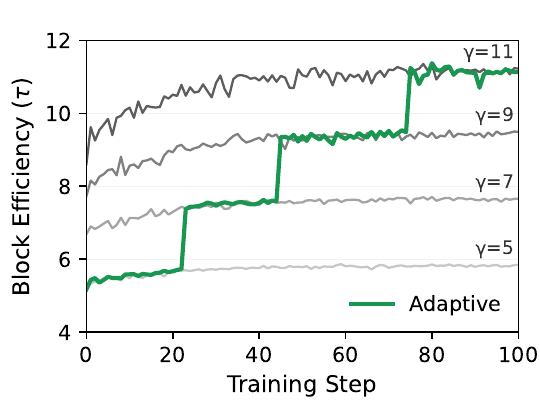}
        \caption{Block efficiency}
        \label{fig:adaptive_gamma_mal}
    \end{subfigure}
    \hfill
    \begin{subfigure}[t]{0.32\linewidth}
        \centering
        \includegraphics[width=\linewidth]{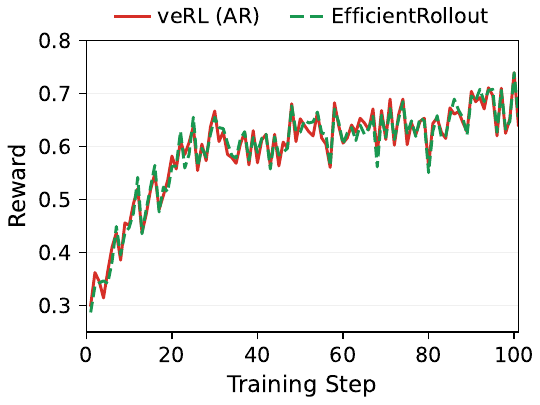}
        \caption{Training reward}
        \label{fig:adaptive_gamma_reward}
    \end{subfigure}

    \caption{
    \textbf{Adaptive draft-length policy and training dynamics on Qwen2.5-7B.}
    (a) Adaptive $\gamma$ reduces rollout-generation time by avoiding overly large drafts early and exploiting longer drafts later.
    (b) The controller raises $\gamma$ as block efficiency $\tau$ improves during training.
    (c) \methodtitle{} follows the \nosd{} reward trajectory, indicating preserved training dynamics.
    }
    \label{fig:adaptive_gamma}
\end{figure}

\textbf{Training dynamics and quality preservation.}
\methodtitle{} accelerates training while preserving training dynamics.
\Cref{fig:adaptive_gamma_reward} shows that \methodtitle{} closely tracks the \nosd{} reward trajectory on Qwen2.5-7B.
This aligns with the lossless nature of SD, which preserves the target distribution~\citep{leviathan2023fast,chen2023accelerating} and is especially important in RL where rollout-distribution shifts can affect training stability~\citep{yu2025dapo,fu2025areal}.
Full reward and validation-accuracy trajectories across all evaluated models are provided in~\cref{app:quality-preservation}.

\textbf{Broader empirical validation.}
We extend the primary evaluation by varying hardware, model scale and parallelism, rollout temperature, and maximum response length.
On H100, it reduces rollout-generation and full-step latency by up to 16.7\% and 12.5\%, respectively.
For Qwen2.5-32B using TP$=2$ on A100, it reduces rollout-generation latency by 13.0\%.
With a lower temperature ($0.6$) and a 16k maximum response length on A100, the reductions reach 40.1\% and 34.0\%, respectively.
Full settings and results are provided in~\cref{app:extended_e2e_evaluation}.

\subsection{Further Analysis of Rollout Acceleration}
\label{subsec:further-analysis}

\textbf{Regime-aware SD toggle policy.}
High $\mal$ alone is insufficient for realized speedup, and SD must be activated only in system-beneficial regimes via the toggle policy $\SDToggle$.
\Cref{fig:toggle_boundary} validates the calibrated roofline boundary of $\SDToggle$ by showing that the predicted speedup correctly separates SD-beneficial and SD-harmful coordinates in the $(B,S)$ plane.
\Cref{fig:toggle_vs_alwayson} isolates the effect of $\SDToggle$ by comparing rollout latency with the same quantized self-drafter when SD is always enabled versus enabled according to $\SDToggle$.
By disabling SD during only the early 6--11\% of decoding steps, the $\SDToggle$ policy yields consistently larger rollout-generation latency reductions relative to AR than always-on SD, despite slightly lower $\mal$.
These gains therefore cannot be attributed to better draft quality; they come from avoiding verification in the initial large-batch regime, where $\VerifyTime/\TargetTime$ is high.
Further A100 and H100 boundary validation and calibration sensitivity for $\SDToggle$ are reported in~\cref{app:toggle_validation}.

\textbf{Adaptive draft-length policy.}
The adaptive $\DraftLength$ policy effectively tracks the latency-reducing $\DraftLength$ as the evolving target policy shifts over training.
We isolate this effect by replacing only the adaptive $\DraftLength$ policy in \methodtitle{} with fixed-$\DraftLength$ variants throughout training.
In this controlled policy ablation, the adaptive $\DraftLength$ policy achieves an overall 19.6\% reduction in rollout-generation latency, compared with 13.5\% and 11.8\% for fixed $\DraftLength{}=\gamma_{\mathrm{low}}=5$ and $\DraftLength{}=\gamma_{\mathrm{high}}=11$, respectively, as shown in~\cref{fig:adaptive_gamma_gen_time}.
This improvement comes from leveraging larger latency reductions at different stages of training, as smaller and larger $\DraftLength$ values are more effective early and later in training, respectively.
Rather than selecting a single fixed $\DraftLength{}$ through post-hoc sweeps, the adaptive $\DraftLength$ policy moves within the pre-specified $\Gamma$ using observed $\mal$ feedback.

\begin{wrapfigure}{r}{0.4\linewidth}
    \vspace{-5mm}
    \centering
    \includegraphics[width=0.97\linewidth]{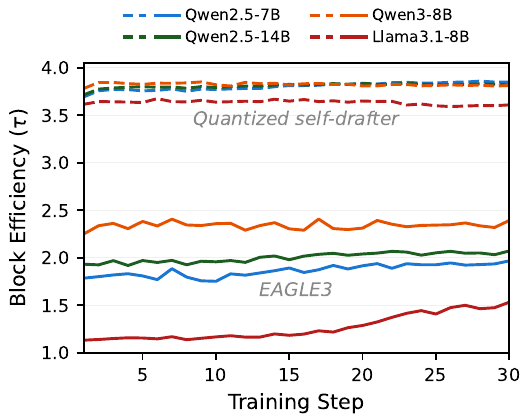}
    \vspace{-1mm}
    \caption{
    \textbf{Block efficiency across pretrained auxiliary drafters.}
    On DAPO-Math-17K, the evaluated pretrained auxiliary drafters achieve lower block efficiency than quantized self-drafters.
    }
    \label{fig:learned_auxiliary_mal_difficulty}
    \vspace{-4mm}
\end{wrapfigure}

\textbf{Learned auxiliary drafting in RL workloads.}
The lower $\mal$ and $\alpha$ of \learned{} in~\cref{tab:e2e_speedup} reflect the practical difficulty of obtaining an auxiliary drafter aligned with long, high-temperature RL rollouts.
Under the prior RL-SD setup of \citet{iso2026accelerating}, we further evaluate auxiliary drafters on DAPO-Math-17K~\citep{yu2025dapo}, the rollout workload used in that setup.
As shown in~\cref{fig:learned_auxiliary_mal_difficulty}, the evaluated auxiliary drafters~\citep{redhatai2025llama31eagle3,redhatai2025qwen3eagle3thinking} achieve only $\mal=1.2$--$2.4$ over the first 30 steps, whereas the quantized self-drafter consistently reaches $\mal=3.6$--$3.9$ under the same setting.
These results suggest that directly reusing public checkpoints or drafters pretrained with fixed-target LLM-serving recipes~\citep{li2025eagle,zhang2025fastgrpo} is insufficient to reliably obtain high $\mal$ in RL rollouts.
Realizing effective speedups with \learned{} may require a more specialized auxiliary-drafter training pipeline~\citep{zhou2024distillspec, iso2026accelerating}, such as collecting in-distribution rollouts, training drafters on target-generated synthetic data, and possibly using more aggressive online adaptation.
Detailed evaluation settings and analyses with additional auxiliary-drafter checkpoints are presented in~\cref{app:workload-matched-auxiliary-drafters}.

\section{Discussion and Future Directions}
\label{sec:discussion_future}

We present \methodtitle{}, a system-aware self-SD framework for RL rollouts, designed around evolving policies, shrinking-batch dynamics, and rollout-tail latency.
By combining a target-induced quantized drafter, dynamic SD toggling, and adaptive draft budgets, \methodtitle{} reduces rollout and end-to-end latency without separate drafter pretraining, warm-up, online adaptation, or invasive RL pipeline changes.

\paragraph{Future directions.}

Several directions remain for improving and extending system-aware self-SD for RL rollouts.
First, we focus on chain verification because it is practical under dynamic batch sizes~\citep{liu2025speculative}; tree verification is orthogonal and could be incorporated into our framework~\citep{li2024eagle,li2025eagle}.
Second, naive RTN may not yield strong early-stage drafters for all model families, motivating less lossy quantization methods that improve drafter quality without introducing excessive per-step overhead.
Adaptive, less-frequent drafter refreshes may further reduce preparation overhead while preserving policy alignment.
Finally, when KV-cache loading dominates in long-context or dense-attention regimes~\citep{cordonnier2020multi}, sparse-attention drafting may further reduce quantized self-SD cost.

\section*{Acknowledgments}

This work was supported by Institute for Information \& communications Technology Promotion (IITP) grant funded by the Korea government (MSIT) (No. 04-26-03-0081, Energy-Efficient Training–Inference System Optimization for Reinforcement Learning-Based Post-Training). This work was also supported by the "Advanced GPU Utilization Support Program" funded by the Government of the Republic of Korea (Ministry of Science and ICT). We thank all members of the FuriosaAI team for their support, with special thanks to Hanjoon Kim and June Paik for their vision and commitment to research.

\bibliographystyle{plainnat}
\bibliography{neurips_2026}


\appendix
\clearpage
{\LARGE \textbf{Appendix}} \par 
\startcontents[sections]
\printcontents[sections]{ }{1}{}

\clearpage

\section{Extended Analysis of Shrinking-Batch Dynamics}
\label{app:long_tail_analysis}

\begin{figure}[t]
    \centering

    \begin{subfigure}[t]{0.32\linewidth}
        \centering
        \includegraphics[width=\linewidth]{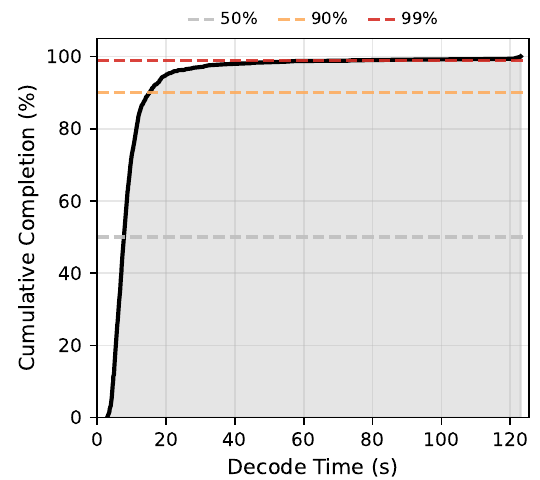}
        \caption{}
        \label{fig:rollout_llama_bottleneck_completion}
    \end{subfigure}
    \hfill
    \begin{subfigure}[t]{0.32\linewidth}
        \centering
        \includegraphics[width=\linewidth]{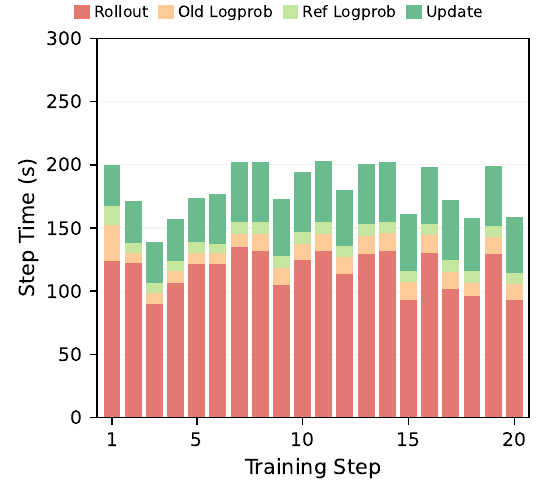}
        \caption{}
        \label{fig:rollout_qwen_bottleneck_step}
    \end{subfigure}
    \hfill
    \begin{subfigure}[t]{0.32\linewidth}
        \centering
        \includegraphics[width=\linewidth]{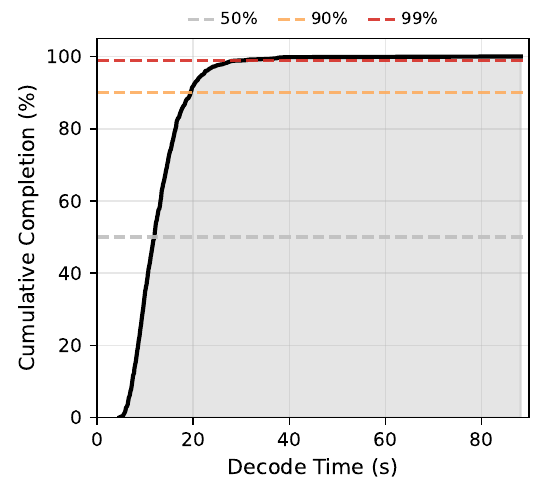}
        \caption{}
        \label{fig:rollout_qwen_bottleneck_completion}
    \end{subfigure}
    \caption{
    Extended rollout-tail analysis.
    (a) Cumulative request-completion curve for Llama3.1-8B-Instruct at the first step of the second epoch; the corresponding step-time decomposition is shown in \cref{fig:rollout_llama_bottleneck_step}.
    (b) Step-time decomposition over the first 20 training steps for Qwen2.5-7B.
    (c) Cumulative request-completion curve for Qwen2.5-7B at the first step of the second epoch.
    }
    \label{fig:rollout_bottleneck_appendix}
\end{figure}

\Cref{fig:rollout_bottleneck_appendix} provides additional evidence that rollout generation remains the dominant bottleneck and exhibits a long-tail completion pattern across models.
For Qwen2.5-7B, \cref{fig:rollout_qwen_bottleneck_step} shows the step-time decomposition over the first 20 training steps.
Although the total step time is shorter than that of Llama3.1-8B-Instruct, rollout is still the largest component.
Across the first 20 steps, rollout accounts for 58.0--71.4\% of the measured four-phase step time, with a mean of 64.3\%.
This confirms that rollout generation is the primary systems bottleneck for Qwen2.5-7B as well.

The completion curves show why rollout dominates wall-clock time.
For Llama3.1-8B-Instruct, \cref{fig:rollout_llama_bottleneck_completion} shows a pronounced long tail at the first step of the second epoch.
Roughly 50\% of requests finish within 10\,s, 90\% within 16\,s, and 99\% within 30\,s, but the final request completes only after about 123\,s.
Thus, a small number of long generations determine the rollout makespan.
Qwen2.5-7B exhibits the same mechanism, although with a milder tail.
As shown in \cref{fig:rollout_qwen_bottleneck_completion}, about 50\% of requests finish within 13\,s, 90\% within 22\,s, and 99\% within 30\,s, while the total rollout makespan is about 88\,s.
Compared with Llama3.1-8B-Instruct, the Qwen tail is shorter, but the final few requests still dominate the end of rollout generation.
Once most requests have completed, the active batch size sharply decreases, leaving a shrinking-batch tail where direct acceleration of the remaining decode steps is most valuable.

\section{System Rationale for Weight-Quantized Self-Drafting in RL Rollouts}
\label{app:why_quantized_sd}

\begin{table*}[t]
\centering
\small
\setlength{\tabcolsep}{8pt}
\renewcommand{\arraystretch}{1.12}
\caption{
Simulation-based single-token decode-time breakdown for Qwen2.5-7B, Qwen2.5-14B, Llama3.1-8B-Instruct at different sequence lengths and batch sizes representing RL rollout tail regimes.
Results assume a single A100-80GiB SXM GPU, FP16 inference, no tensor parallelism, and full attention.
Bold entries indicate weight-loading costs targeted by quantization; sparse attention targets the attention term.
}
\label{tab:decode_breakdown_qwen_llama}
\resizebox{\textwidth}{!}{%
\begin{tabular}{ccccccc}
\toprule
\textbf{Model}
& \textbf{(Seq, Batch)}
& \textbf{FFN}
& \textbf{QKVO Proj}
& \textbf{Attn ($QK \cdot V$)}
& \textbf{LM Head}
& \textbf{Nonlinear} \\
\midrule
\multirow{4}{*}{Qwen2.5-7B}
& (2k, 8) & \textbf{75.2\%} & \textbf{10.9\%} & 6.2\% & 7.2\% & 0.5\% \\
& (2k, 4) & \textbf{77.8\%} & \textbf{11.2\%} & 3.2\% & 7.4\% & 0.3\% \\
& (4k, 2) & \textbf{77.9\%} & \textbf{11.2\%} & 3.2\% & 7.4\% & 0.2\% \\
& (8k, 1) & \textbf{78.0\%} & \textbf{11.2\%} & 3.2\% & 7.5\% & 0.1\% \\
\midrule

\multirow{4}{*}{Qwen2.5-14B}
& (2k, 8) & \textbf{65.0\%} & \textbf{19.3\%} & 10.3\% & 5.0\% & 0.4\% \\
& (2k, 4) & \textbf{68.7\%} & \textbf{20.4\%} & 5.4\% & 5.2\% & 0.2\% \\
& (4k, 2) & \textbf{68.8\%} & \textbf{20.4\%} & 5.4\% & 5.3\% & 0.1\% \\
& (8k, 1) & \textbf{68.8\%} & \textbf{20.4\%} & 5.4\% & 5.3\% & 0.1\% \\
\midrule
\multirow{4}{*}{Llama3.1-8B}
& (2k, 8) & \textbf{65.4\%} & \textbf{15.6\%} & 12.4\% & 6.1\% & 0.5\% \\
& (2k, 4) & \textbf{69.9\%} & \textbf{16.7\%} & 6.7\% & 6.5\% & 0.3\% \\
& (4k, 2) & \textbf{70.0\%} & \textbf{16.7\%} & 6.7\% & 6.5\% & 0.2\% \\
& (8k, 1) & \textbf{70.0\%} & \textbf{16.7\%} & 6.7\% & 6.5\% & 0.1\% \\
\bottomrule
\end{tabular}%
}
\end{table*}

\subsection{Detailed Decode-Time Decomposition in Rollout-Tail Regimes}
\label{app:decode_breakdown_more}

We follow the roofline-based simulation methodology of prior work~\cite{kim2025beyond} and decompose the single-token decode cost for representative rollout-tail regimes, where each component latency is estimated as the maximum of its compute time and memory-transfer time.
\Cref{tab:decode_breakdown_qwen_llama} reports the resulting breakdown for Qwen2.5-7B/14B~\cite{Yang2024Qwen25TR} and Llama3.1-8B-Instruct~\cite{grattafiori2024llama}.

The same qualitative trend holds across the evaluated models.
In shrinking-batch rollout regimes, FFN and QKVO projection together dominate the decode cost, while attention contributes only a relatively small fraction.
These additional results further support the design choice of quantization-based self-SD, which directly reduces the dominant linear component, rather than sparse-attention drafting, which mainly targets the much smaller attention component.
We do not consider KV-cache quantization in this work, although it could become useful in much longer rollout settings, e.g., beyond 64k tokens, where attention cost may again become significant.

\subsection{Practical Trade-offs between Quantized and Sparse-Attention Self-Drafting}
\label{app:practical_quantized_vs_sparse}

From a practical systems perspective, sparse-attention drafting interacts poorly with the serving engine.
To achieve high acceptance, sparse-attention self-SD typically requires token-wise sparsity patterns that are updated dynamically at each draft step~\cite{tang2024quest,yue2026specattn}.
However, such fine-grained sparsity is difficult to realize efficiently in vLLM, whose memory management is page-granular rather than token-granular~\cite{kwon2023efficient}.
If we instead use vLLM-friendly sparse patterns, such as page-level sparsity or sliding-window attention, acceptance drops noticeably.
We implemented a sparse-attention self-SD method following prior work~\cite{yue2026specattn}.
Under sparsity budgets that are cheap enough to yield speedup in practice, e.g., retaining 256 tokens, the resulting drafter remains relatively weak: with $\gamma=5$, block efficiency $\tau$ typically stays around 3--4.
In contrast, a 4-bit weight-quantized drafter works well in our regime.
Even with the simplest asymmetric RTN (round-to-nearest) quantization, we achieve block efficiency $\tau=5.2$ at $\gamma=5$.
This design also naturally synergizes with vLLM through optimized W4A16 Marlin kernels~\cite{frantar2025marlin}.
Taken together, these observations make quantized self-SD the most effective and practical self-drafting strategy for RL rollout acceleration in our setting.

\subsection{W4 versus W8: Latency--Acceptance Trade-off}
\label{app:w4_vs_w8}

\begin{table}[t]
\centering
\small
\setlength{\tabcolsep}{6pt}
\renewcommand{\arraystretch}{1.10}
\caption{
W4--W8 latency--acceptance trade-off.
We report $\widehat{\DraftTime}/\widehat{\TargetTime}$ predicted by our roofline model under memory-bound, zero-overhead assumptions, and block efficiency $\tau$ for $\gamma \in \{3,5,7\}$ measured on Qwen2.5-7B-Instruct at batch size 1 and sequence length 2k.
}
\label{tab:w4_vs_w8}
\vspace{1mm}
\begin{tabular}{lcccc}
\toprule
\textbf{Drafter}
&
\makecell{\textbf{Idealized}\\$\widehat{\DraftTime}/\widehat{\TargetTime}$}
&
\makecell{\boldmath$\tau$\\\textbf{($\gamma=3$)}}
&
\makecell{\boldmath$\tau$\\\textbf{($\gamma=5$)}}
&
\makecell{\boldmath$\tau$\\\textbf{($\gamma=7$)}} \\
\midrule
RTN W8 & 0.573 & 3.94 & 5.87 & 7.79 \\
RTN W4 & 0.360 & 3.59 & 5.18 & 6.70 \\
\bottomrule
\end{tabular}
\end{table}

\Cref{tab:w4_vs_w8} compares W4 and W8 RTN quantized drafters.
As expected, W8 provides higher drafting quality and achieves larger block efficiency across draft lengths.
However, W4 remains reasonably accurate while making the draft path substantially cheaper: in the idealized memory-bound model, W4 reduces the draft-to-target latency ratio to 0.360, compared with 0.573 for W8.
This difference is important because SD speedup depends not only on accepted length, but also on whether the drafter is cheap enough to amortize target verification overhead.

W8 also introduces a larger memory footprint in the rollout engine.
For example, its drafter weights require roughly twice the memory of W4, reaching about 16.3\,GiB for the Qwen2.5-14B drafter.
This additional memory pressure is undesirable during RL rollouts, where persistent training weights, rollout weights, and vLLM KV-cache memory already compete for GPU capacity.
Moreover, although W8 improves block efficiency, its draft path is not sufficiently cheaper than target AR decoding to reliably translate the higher acceptance into end-to-end speedup when considering realistic overheads.

\begin{figure}[t]
    \centering
    \includegraphics[width=0.5\linewidth]{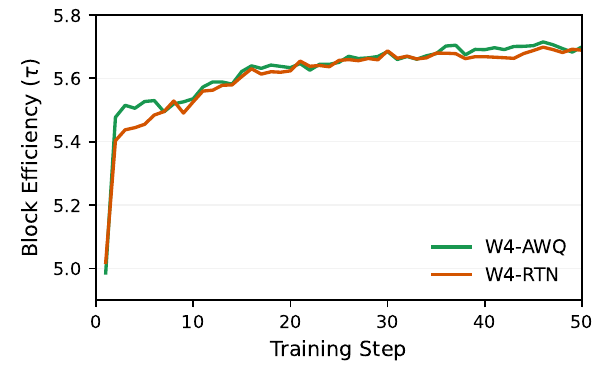}
    \caption{
    Block efficiency $\tau$ over the first 50 training steps on Qwen2.5-7B with W4-RTN and W4-AWQ drafters.
    }
    \label{fig:appendix_quant_mal}
\end{figure}

\subsection{Choosing RTN for Step-Wise Drafter Refresh}
\label{app:awq_quantization}

We also experimented with an activation-aware quantization (AWQ) pipeline~\cite{lin2024awq} for constructing the W4 drafter. We collect hidden-state statistics from the previous training step and use them to perform AWQ quantization, with the goal of improving drafter quality over naive round-to-nearest (RTN) quantization.
\Cref{fig:appendix_quant_mal} compares the block efficiency $\tau$ achieved by W4-RTN and W4-AWQ on Qwen2.5-7B over the first 50 training steps.
W4-AWQ provides a small advantage during the first few steps, but the gap quickly disappears as the RTN drafter quality improves and saturates.
This suggests that, for Qwen2.5-7B, simple RTN quantization is already sufficient for maintaining a high-quality self-drafter during RL training.

The main drawback of AWQ-style quantization is its additional overhead.
Unlike RTN, activation-aware methods require collecting calibration data and running a more expensive quantization procedure before each drafter refresh.
Since \methodtitle{} refreshes the drafter every training step, this overhead can become a significant burden unless the quantization pipeline is heavily optimized.
Nevertheless, less lossy quantization may be useful for models whose RTN-quantized drafter has low initial acceptance.
We leave more advanced step-wise quantization schemes as future work.

We also do not use bitsandbytes (BNB) 4-bit quantization~\cite{dettmers2023qlora}.
Although BNB-style 4-bit quantization can provide more calibrated low-bit weights, its execution path is not as well optimized for production-grade vLLM rollout inference as the AWQ-compatible W4A16 Marlin kernels used in our implementation.

\section{Policy Sharpening Improves Quantized Drafter Alignment}
\label{app:entropy_acceptance}

\begin{figure}[t]
    \centering
    \includegraphics[width=\textwidth]{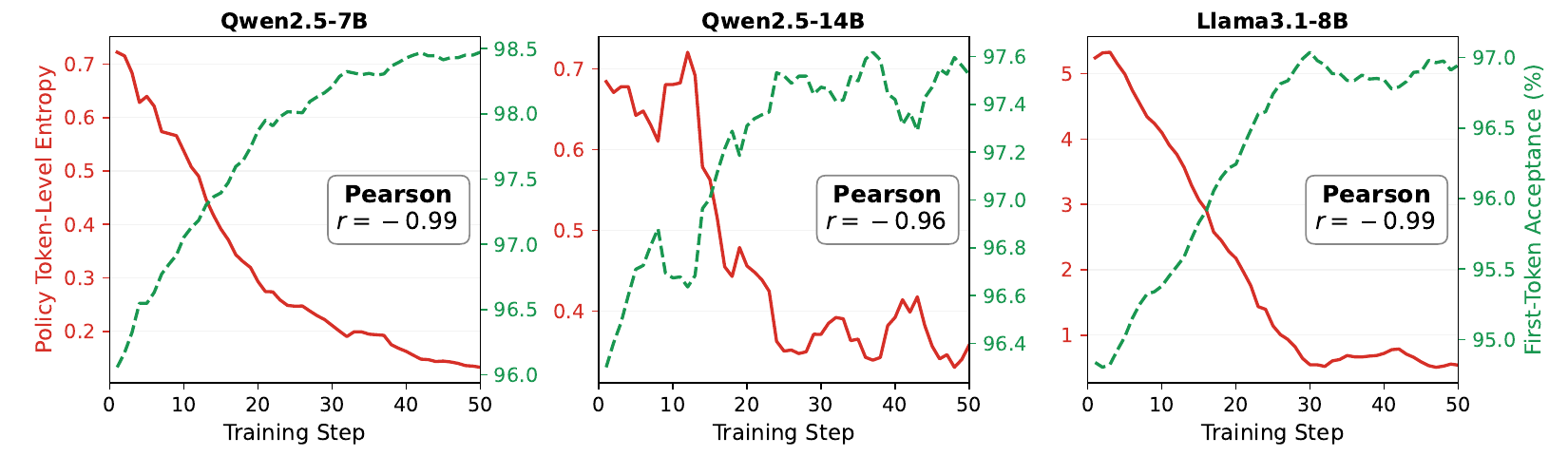}
    \caption{
    Policy sharpening improves quantized drafter alignment.
    As target-policy entropy decreases, first-token acceptance of the RTN W4 drafter increases across models.
    }
    \label{fig:appendix_entropy_acceptance}
\end{figure}

\begin{figure}[t]
    \centering
    \includegraphics[width=\textwidth]{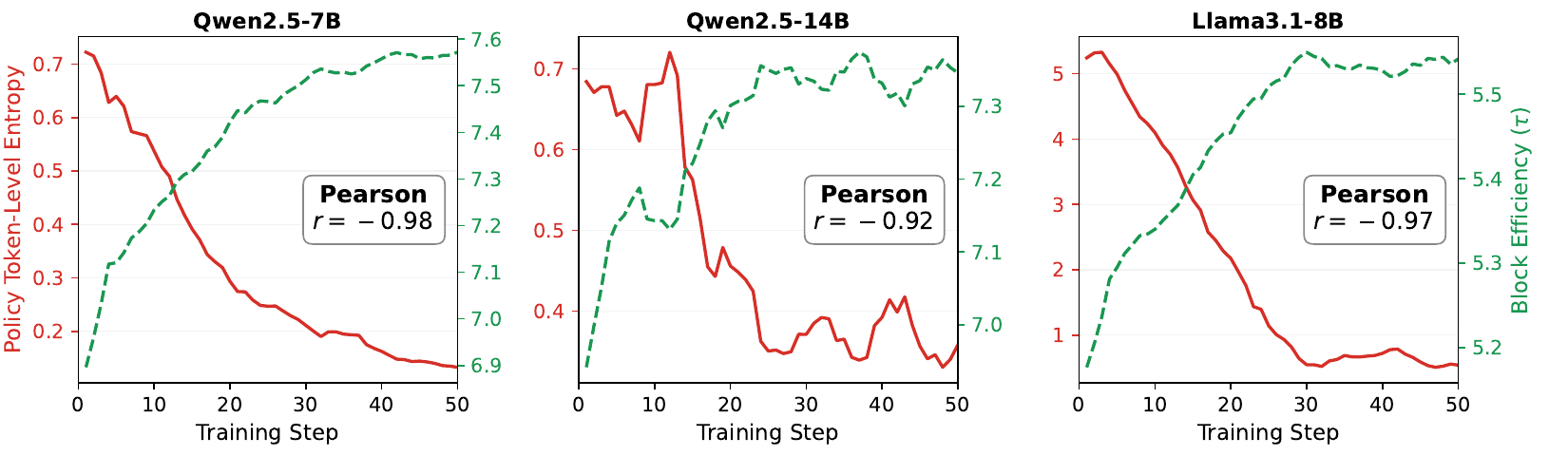}
    \caption{
    Policy sharpening improves quantized drafter alignment.
    As target-policy entropy decreases, block efficiency $\tau$ of the RTN W4 drafter increases across models.
    }
    \label{fig:appendix_entropy_mal}
\end{figure}

We analyze how RL training affects quantized self-drafting under a fixed SD configuration, without the regime-aware toggle or adaptive draft-length control.
At each training step, we construct the drafter by applying vanilla round-to-nearest (RTN) 4-bit quantization to the current target model, and use this quantized copy to propose drafts with a fixed draft length.
We measure two quantities: first-token acceptance rate, defined as the fraction of SD iterations in which the first drafted token is accepted by the full-precision target model, and block efficiency $\tau$, defined as the average number of target-distributed tokens produced per SD iteration.
The former directly reflects local drafter--target alignment at the current prefix, while the latter captures the resulting effectiveness of each SD block.

\Cref{fig:appendix_entropy_acceptance} shows the target-policy entropy and first-token acceptance rate over training for Qwen2.5-7B, Qwen2.5-14B, and Llama3.1-8B-Instruct.
Across all three models, RL training reduces policy entropy while first-token acceptance steadily increases, with strong negative Pearson correlations ranging from $-0.96$ to $-0.99$.
\Cref{fig:appendix_entropy_mal} shows that block efficiency follows the same trend, with Pearson correlations ranging from $-0.92$ to $-0.98$.
These results suggest that RL post-training sharpens the target distribution, making small quantization-induced perturbations less likely to change accepted draft tokens and thereby improving block efficiency over training.

\section{Implementation and Calibration Details of \methodtitle{}}
\label{app:method-details}

\begin{figure}[t]
    \centering
    \includegraphics[width=0.6\linewidth]{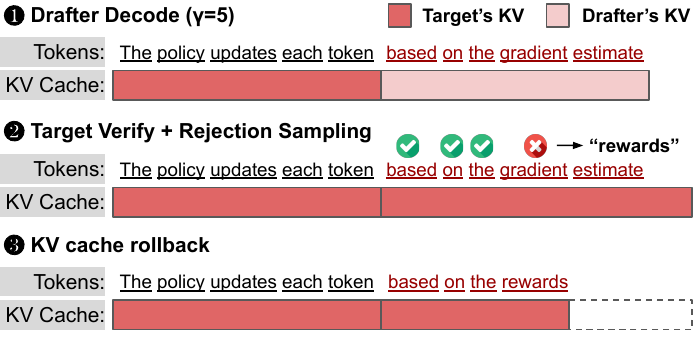}
    \caption{
    Quantized self-SD with shared KV cache, target verification via rejection sampling, and rollback of rejected KV cache.
    }
    \label{fig:appendix_shared_kv}
\end{figure}

\subsection{KV-Cache Sharing between the Quantized Drafter and Target}
\label{app:shared-kv-sd}

\Cref{fig:appendix_shared_kv} illustrates the shared-KV execution pipeline used by \methodtitle{}.
The quantized drafter shares the target model's KV-cache storage, so drafting does not require an additional KV cache for the draft model.
At the start of each SD iteration, the prefix KV cache consists only of clean states produced by the full-precision target model.
The W4 drafter then proposes $\gamma$ tokens using quantized weights and writes provisional KV states for the drafted positions into the shared KV cache.

The target model verifies the drafted tokens in parallel using full-precision weights.
During verification, the target overwrites the provisional KV entries at the verified positions with clean target-generated KV states and performs rejection sampling to preserve the target-model sampling distribution.
If all drafted tokens are accepted, the target additionally samples the bonus token and the next SD iteration continues from the updated clean KV cache.
If a drafted token is rejected, the target resamples at the first rejected position from the adjusted target distribution, and all KV states after that position are discarded.
Thus, every new SD iteration begins from a prefix whose KV states are generated by the full-precision target model.

This design has two practical advantages for RL rollouts.
First, sharing the KV cache avoids the memory overhead of maintaining a separate drafter KV cache, which is important when rollout generation uses long maximum response lengths.
Second, because the drafter is reconstructed by quantizing the current target model at each training step, it remains synchronized with the evolving RL policy without auxiliary drafter training or online adaptation.

\subsection{Roofline-Based SD Toggle Model and Calibration}
\label{app:toggle_calibration}

We provide additional details on the parameters used in the SD toggle model and on the calibration procedure.

\paragraph{Parameter details.}
$W_T$ denotes the weight size of the target model in bytes, and $W_D$ denotes the weight size of the W4 drafter in bytes.
Because only the FFN and QKVO projection layers are quantized to 4 bits, while the embedding layer, LM head, and normalization layers remain in higher precision, the ratio $W_D/W_T$ is not exactly $25\%$; in practice, it is about $33$--$36\%$ of the original model size.
$C_{\mathrm{dense}}$ is the per-token dense compute cost, including FFN, QKVO projection, and LM-head computation, and $C_{\mathrm{attn}}$ is the per-token attention compute cost.
These quantities can be determined statically by the model architecture.

The effective per-token KV-cache traffic is modeled as
\[
\kappa_{\mathrm{eff}} = \rho_{\kappa}\,\kappa_{\mathrm{theoretical}},
\]
where $\kappa_{\mathrm{theoretical}}$ is the theoretical KV traffic determined by the model configuration,
\[
\kappa_{\mathrm{theoretical}}
=
(\#\mathrm{layers})
\times 2
\times (\#\mathrm{KV\ heads})
\times (\mathrm{head\ dim})
\times 2~\mathrm{bytes},
\]
with the factor of $2$ for key/value tensors and the final $2$ bytes corresponding to FP16 precision.
This quantity can be computed statically from the model architecture.
The scaling factor $\rho_{\kappa}$ is typically smaller than $1$, reflecting effects such as cache reuse and overlap with other operations.

The factor $\eta_D$ captures the practical overhead of W4A16 drafting.
Although weight quantization reduces the raw model weight size substantially, this does not translate directly into proportional memory-time reduction, because quantized decoding still incurs additional costs such as on-the-fly dequantization and kernel-level overhead.
Thus, $\eta_D$ models the aggregate overhead beyond the idealized weight-byte reduction.

Finally, we model the residual overhead terms as $c_T B$, $c_D B$, and $c_V B$ for target decoding, drafter decoding, and verification, respectively.
Empirically, we found that these residual overheads are dominated by batch-dependent effects, making a linear-in-batch approximation sufficiently accurate.

For TP$>1$, the target and W4 drafter are co-sharded under the same TP layout, and each rank requantizes only its local target shard, so drafter refresh requires no additional model-wide collective.
We can account for TP communication by adding the measured collective cost $c_{\mathrm{comm}}$ to each calibrated latency estimate:
\begin{equation}
\widehat{T}_{x,\mathrm{TP}}(B,S,\gamma)
=
\widehat{T}_x(B,S,\gamma)
+
c_{\mathrm{comm}},
\qquad
x\in\{p,q,V\}
\label{eq:tp_aware_roofline}
\end{equation}
The adjusted target, drafter, and verification costs enter the same toggle rule in~\cref{eq:sd_speedup_toggle}; end-to-end TP$=2$ results are reported in~\cref{app:extended_e2e_evaluation}.

\paragraph{Calibration procedure.}
We calibrate the model in three stages.
First, we measure the effective compute throughput $\mathrm{F}_{\mathrm{eff}}$ using a hardware-specific microbenchmark based on repeated dense matrix multiplications.
Second, for each model of interest, we sweep full-precision target decoding, verification, and W4 quantized drafter decoding over a range of batch sizes and sequence lengths, with draft lengths $\gamma \in \{3, 7, 11, 15\}$.
Third, we fit the remaining parameters in two steps.
We first fit $\mathrm{BW}_{\mathrm{eff}}$, $\rho_{\kappa}$, and $c_T$ jointly using the target-model sweep data, and then fix them.
We next fit $\eta_D$, $c_D$, and $c_V$ using the drafter and verification sweep data.
Because self-SD uses the same architecture for the target and drafter, the compute term $C(B,S)$ is shared between target decoding and drafter decoding.
All fitted parameters are independent of $\gamma$; the dependence on draft length enters only through the verification compute term and the speedup model in \cref{eq:sd_speedup}.
This allows a model calibrated with a small set of draft lengths to generalize across other draft lengths without re-calibration.

In principle, $\mathrm{BW}_{\mathrm{eff}}$ should depend only on the hardware platform.
However, unlike $\mathrm{F}_{\mathrm{eff}}$, it is difficult to benchmark memory bandwidth in a directly comparable standalone form for our decoding workloads.
We therefore fit $\mathrm{BW}_{\mathrm{eff}}$ separately for each model.
In practice, the fitted values are consistent across models, typically falling in the range of 1560--1620~GB/s.

\begin{figure*}[t]
    \centering
    \includegraphics[width=0.8\textwidth]{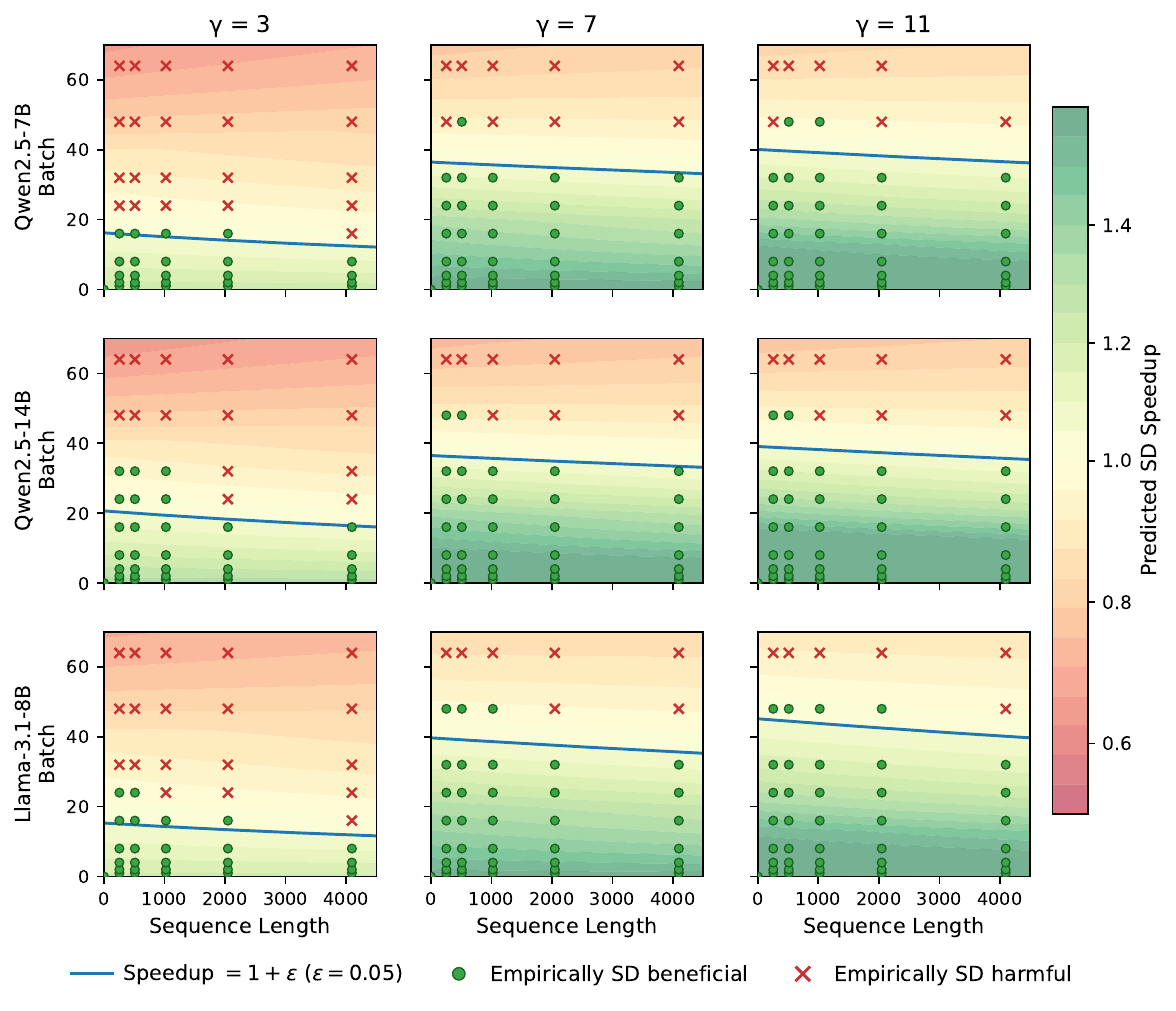}
    \vspace{-2pt}
    \caption{
    Validation of the roofline-based SD toggle policy.
    Background colors show predicted speedup, the blue line marks $\mathrm{Speedup}_{\mathrm{SD}}=1+\epsilon$, and markers indicate empirical SD-beneficial/harmful points.
    }
    \label{fig:appendix_sd_grid}
\end{figure*}

\subsection{Validation of the Roofline-Based SD Toggle Policy}
\label{app:toggle_validation}

\paragraph{Toggle-boundary validation.}
We validate the calibrated roofline model used by the SD toggle policy against measured W4 self-SD component timings.
For each model, we sweep active batch size $B$, sequence length $S$, and draft length $\DraftLength$, and measure $\TargetTimebs$, $\DraftTimebs$, and $\VerifyTimebsg$ on a single A100 GPU with TP$=1$.
Using these measured components, we compute the empirical SD speedup using the same notation as~\cref{eq:sd_speedup}:
\begin{equation}
\mathrm{Speedup}_{\mathrm{emp}}(B,S,\gamma)
=
\frac{\mal \TargetTimebs}
{\DraftLength \DraftTimebs + \VerifyTimebsg}.
\label{eq:empirical_sd_speedup}
\end{equation}
For this validation, we use the optimistic setting $\mal=\DraftLength$ to isolate the cost-model accuracy from block-efficiency variation.
\Cref{fig:appendix_sd_grid} compares the policy-predicted speedup surface with empirical measurements.
The background color shows the speedup predicted by the calibrated roofline model, and the blue contour indicates the toggle boundary at $\mathrm{Speedup}_{\mathrm{SD}}=1+\epsilon$ with $\epsilon=0.05$.
Green circles denote measured points where SD is empirically beneficial, while red crosses denote points where SD is harmful.
Across models and draft lengths, the predicted boundary closely tracks the empirical beneficial/harmful frontier.
The policy is slightly conservative because of the safety margin $\epsilon$: several empirically beneficial points lie just above the blue boundary, but the policy intentionally keeps SD disabled unless it predicts a sufficient margin.

\begin{table*}[t]
\centering
\small
\setlength{\tabcolsep}{10pt}
\renewcommand{\arraystretch}{1.08}
\caption{Sensitivity of the Roofline toggle policy to independent $\pm10\%$ perturbations of each calibrated term. $\Delta$ is the worst absolute change in sign accuracy relative to the unperturbed calibration; flips are the fraction of grid cells whose AR/SD decision changes.}
\label{tab:toggle_calibration_sensitivity}
\vspace{-4pt}
\begin{tabular}{lcccc}
\toprule
\textbf{Perturbed Term}
& \textbf{A100 $\Delta$ (pp)}
& \textbf{A100 Flips (\%)}
& \textbf{H100 $\Delta$ (pp)}
& \textbf{H100 Flips (\%)} \\
\midrule
$\mathrm{BW}_{\mathrm{eff}}$ & 2.7 & 4.9 & 2.2 & 5.1 \\
$\mathrm{F}_{\mathrm{eff}}$  & 2.7 & 2.2 & 1.7 & 2.8 \\
$\eta_D$                     & 4.9 & 6.6 & 7.3 & 16.9 \\
$\kappa_{\mathrm{eff}}$     & 0.0 & 0.0 & 0.0 & 0.0 \\
$c_T$                         & 2.7 & 3.3 & 2.2 & 2.2 \\
$c_D$                         & 2.2 & 4.4 & 3.4 & 4.5 \\
$c_V$                         & 0.5 & 2.2 & 1.7 & 0.6 \\
\bottomrule
\end{tabular}
\end{table*}

\paragraph{Calibration sensitivity.}
We independently perturb each of the seven calibrated terms by $\pm10\%$ and compare the resulting policy decisions with those from the unperturbed calibration.
The two metrics capture complementary effects: $\Delta$ measures the largest absolute change in aggregate sign accuracy, whereas flips count grid cells whose AR/SD decision changes.
Across the A100 and H100 validation grids, six of the seven terms change sign accuracy by at most 3.4 pp and flip at most 5.1\% of decisions, as shown in~\cref{tab:toggle_calibration_sensitivity}.
The W4 drafter-cost multiplier $\eta_D$ is the most sensitive term, flipping 6.6\% and 16.9\% of decisions on A100 and H100, respectively.
However, an independent calibration on a second A100 node changes $\eta_D$ by only $-0.5\%$ to $+0.9\%$ across the three models.
This cross-node variation is far smaller than the $\pm10\%$ perturbation range, supporting the practical stability of calibration.

\begin{table}[t]
\centering
\small
\setlength{\tabcolsep}{7pt}
\renewcommand{\arraystretch}{1.08}
\caption{H100 toggle-policy validation after one-time model--hardware recalibration.}
\label{tab:h100_toggle_validation}
\vspace{4pt}
\begin{tabular}{lcc}
\toprule
\textbf{Model}
& \textbf{Sign Accuracy (\%)}
& \textbf{Harmful Activation (\%)} \\
\midrule
Qwen2.5-7B  & 88.6 & 4.0 \\
Qwen2.5-14B & 94.5 & 0.0 \\
Llama3.1-8B & 91.6 & 0.6 \\
\bottomrule
\end{tabular}
\end{table}

\paragraph{Cross-hardware H100 validation.}
We repeat the boundary validation on H100 after recalibrating once for each model--hardware pair.
Sign accuracy is the fraction of grid cells for which the predicted SD-beneficial or SD-harmful label matches measurement; harmful activation is the fraction of cells activated with $\epsilon=0.05$ for which measured SD is slower than AR.
As shown in~\cref{tab:h100_toggle_validation}, sign accuracy ranges from 88.6\% to 94.5\%, while harmful activation remains between 0.0\% and 4.0\%.
These results show that the same calibration-and-control procedure recovers a reliable hardware-specific boundary on H100, rather than assuming a universal boundary across hardware.

\section{Detailed Experimental Setup}
\label{app:experimental_setup}

\subsection{Infrastructure, Datasets, and Reporting Window}
\label{app:infra_dataset_reporting}

We use veRL~\cite{sheng2025hybridflow} v0.7.0 with vLLM~\cite{kwon2023efficient} v0.11.2.
The training pipeline uses Ray for distributed execution, FSDP for actor training, and vLLM as the rollout backend.
All experiments are conducted on a single node with 8 NVIDIA A100-SXM4-80GB GPUs.
We use the SimpleRL math datasets~\cite{zeng2025simplerl}.
SimpleRL-8k-hard and SimpleRL-8k-medium correspond to MATH~\cite{hendrycks2021measuring} Level 3--5 and Level 1--4 problems, respectively.
Following the SimpleRL recipe, we use SimpleRL-8k-hard for the Qwen models and easier one for Llama3.1-8B-Instruct, reflecting the stronger reasoning capability of the Qwen base models in this setting~\cite{zeng2025simplerl}.
For timing summaries, we exclude the first training step because it includes one-time distributed setup and CUDA-graph initialization overheads that do not repeat in steady-state training.
All reported averages are computed over the following 100 steps.

\subsection{Metric Measurement Details}
\label{app:metric_measurement}

For all methods, preparation time, rollout-generation time, and step time are measured throughout training and averaged over training steps.
For \methodtitle{}, the draft length can change only at step boundaries; we therefore report $\mal$ and $\bar{\gamma}$ averaged over training steps, so $\bar{\gamma}$ need not be an integer.
We compute $\alpha$ from the measured $\mal$ using the equation $\mal=(1-\alpha^{\DraftLength+1})/(1-\alpha)$.
When $\gamma$ varies over training, we apply the same inversion within fixed-$\gamma$ portions of the trajectory and average the resulting rates weighted by the number of training steps in each portion.
Preparation time includes method-specific overhead required to enable SD.
For \history{}, it includes rollout-history caching and loading overhead; prefix-matching costs incurred during generation are included in Rollout Gen.
For \Learned{}, it includes hidden-state extraction, which incurs negligible overhead in our implementation, and online drafter-training overhead, including drafter forward/backward passes and optimizer updates.
For \alwayssd{} and \methodtitle{}, it includes per-step drafter quantization overhead.

\subsection{Training and Method Configuration}
\label{app:training_method_config}

We use a train batch size of 128 and generate 8 rollouts per prompt, with a maximum rollout length of 8,192 tokens.
Training is performed with a learning rate of $5\times 10^{-7}$ and a mini-batch size of 128.
Because each mini-batch contains samples from the current policy only, training is purely on-policy and does not require policy-ratio clipping.
The default sampling temperature is set to 1.0 with data parallelism (TP=1).
For the Qwen base models, we use a KL loss coefficient of $10^{-4}$ following SimpleRL~\citep{zeng2025simplerl}, and for the Llama instruct model, we use $10^{-2}$ following FastGRPO~\citep{zhang2025fastgrpo}.
For training stability, we further apply token-level rejection sampling [0.5, 2.0] and frequency penalty 0.05 for Qwen2.5-14B.
For \adaptivepolicy{}, we use the same configuration for all models: $\Gamma=\{5,7,9,11\}$, $\alpha_{\mathrm{up}}=0.94$, $\alpha_{\mathrm{down}}=0.85$, and $P=2$.
We omit $\gamma=3$ because it is consistently dominated by, or comparable to, $\gamma=5$ in the rollout regimes encountered during training.

\subsection{Rollout-History-Based Drafting Baseline}
\label{app:history_drafting}

We use Spec-RL~\cite{liu2025spec} as a representative rollout-history-based drafting baseline.
To our knowledge, it is the only publicly available baseline in this category implemented in a veRL+vLLM training stack.
Spec-RL collects generations from previous epochs and, starting from the second epoch, reuses matching prefixes from this rollout history as draft tokens for the current policy.

This design is not strictly target-distribution preserving.
First, after a token is rejected, Spec-RL does not perform rejection sampling from the corrected target distribution.
Exact SD requires resampling from an adjusted distribution at the first rejected position~\cite{leviathan2023fast}, which in turn requires access to the full probability distribution at that position.
In contrast, rollout-history-based drafting makes exact rejection sampling difficult because it stores only past tokens and their corresponding token probabilities, rather than full logits over the vocabulary for every generated position.
Storing full logits for all positions across an epoch would introduce prohibitive memory and storage overhead.
Second, Spec-RL uses a lenience factor of $e^{0.5}$ in its best-performing configuration to increase the length of reused prefixes.
This heuristic makes token acceptance more permissive, but further departs from exact target-distribution-preserving speculative decoding.
We therefore treat Spec-RL as a lossy rollout-history-based baseline.

\begin{figure}[t]
    \centering

    \begin{subfigure}[t]{0.32\linewidth}
        \centering
        \includegraphics[width=\linewidth]{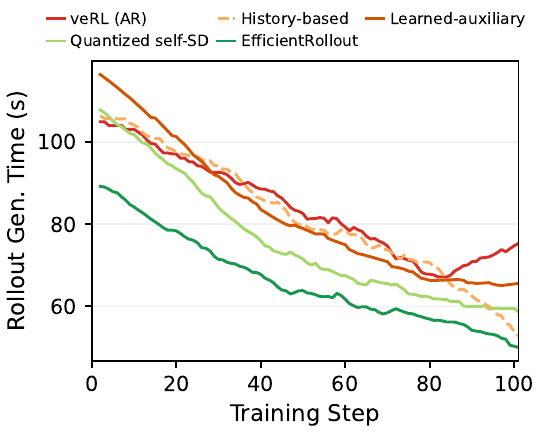}
        \caption{Qwen2.5-7B}
        \label{fig:app_qwen7b_gen_time}
    \end{subfigure}
    \hfill
    \begin{subfigure}[t]{0.32\linewidth}
        \centering
        \includegraphics[width=\linewidth]{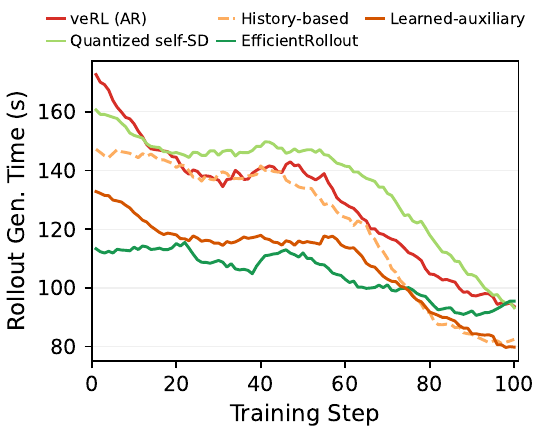}
        \caption{Qwen2.5-14B}
        \label{fig:app_qwen14b_gen_time}
    \end{subfigure}
    \hfill
    \begin{subfigure}[t]{0.32\linewidth}
        \centering
        \includegraphics[width=\linewidth]{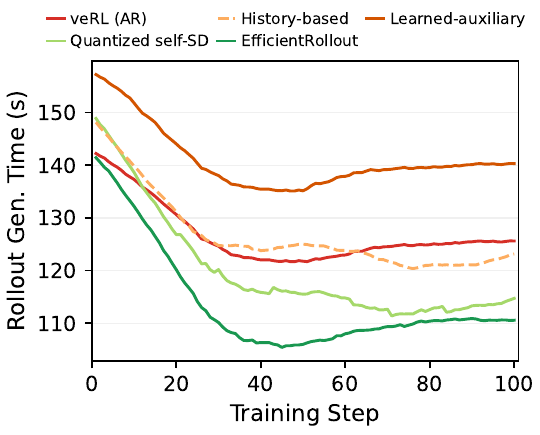}
        \caption{Llama3.1-8B}
        \label{fig:app_llama8b_gen_time}
    \end{subfigure}

    \caption{
    Rollout-generation time over training steps.
    \methodtitle{} consistently reduces rollout time, whereas other SD baselines provide limited gains or can be slower than \nosd{}.
    Curves are smoothed for visualization; \cref{tab:e2e_speedup} reports averages of the unsmoothed measurements.
    }
    \label{fig:app_gen_time_breakdown}
\end{figure}

\subsection{Learned Auxiliary Drafting Baseline}
\label{app:learned_auxiliary_drafting}

Existing RL rollout acceleration systems with \learned{} use different execution stacks.
FastGRPO~\citep{zhang2025fastgrpo} uses a HuggingFace Transformers-based rollout pipeline rather than a serving backend such as vLLM or SGLang.
FastRL~\citep{hu2026taming} uses SGLang~\citep{zheng2024sglang} for rollout generation, while NeMo RL~\citep{iso2026accelerating} uses vLLM with NVIDIA's own post-training framework.
A direct system-level comparison would therefore mix drafter design with backend and framework differences, including batching, KV-cache management, SD execution paths, and training-loop implementation.

To isolate the drafter design, we implement an EAGLE3-style \learned{} baseline in the same veRL+vLLM stack used by the other methods in our paper.
A practical requirement of this baseline is that a compatible drafter must be available for each target model family before RL post-training begins.
For Qwen2.5-7B and Qwen2.5-14B, we did not find publicly available compatible EAGLE3 drafters, so we pretrain the drafters on ShareGPT~\citep{aeala2023sharegpt} using the official vLLM SD training pipeline following EAGLE3~\citep{li2025eagle}.
Each drafter is trained for 10 epochs, and we select the checkpoint with the best validation performance.
For Llama3.1-8B-Instruct, we use the publicly available RedHatAI pretrained EAGLE3 drafter~\citep{redhatai2025llama31eagle3}.

For online adaptation during RL, we carefully follow the rollout-SD design choices of FastRL~\citep{hu2026taming} and NeMo RL~\citep{iso2026accelerating}.
Specifically, we capture hidden states during the actor forward pass in the log-probability computation phase and train the drafter inline immediately after each micro-batch of the actor forward.
This avoids cross-step buffering and CPU offload while allowing the drafter to consume the freshly captured target hidden states.
We follow FastRL's packed layout, next-position shift convention, and combined loss with a hidden-regression term and a soft cross-entropy probability-matching term at the 1:0.1 weight ratio; the drafter is trained with a small fixed AdamW learning rate (1e-6).
As a loss-design check, removing the hidden-regression term while keeping only the soft cross-entropy term (matching the NeMo RL formulation) produced similar online block-efficiency trajectories, so we use the FastRL-style combined loss for the primary baseline.
We use a separate AdamW optimizer for the drafter and report the direct drafter-update wall time as method-specific preparation time.

We do not implement FastRL's asynchronous background drafter training, CPU offload, or overlap optimizations.
These techniques introduce additional system-level design choices beyond the drafter itself and would make the comparison depend on overlap efficiency rather than only rollout-time SD behavior.
For decoding, we use chain drafting with a fixed draft length $\gamma=3$, following the best-performing setting reported by~\citet{iso2026accelerating}.

\begin{figure}[t]
    \centering

    \begin{subfigure}[t]{0.32\linewidth}
        \centering
        \includegraphics[width=\linewidth]{figures/appendix_fidelity_reward_qwen7b.pdf}
        \caption{Qwen2.5-7B}
        \label{fig:appendix_qwen7b_reward}
    \end{subfigure}
    \hfill
    \begin{subfigure}[t]{0.32\linewidth}
        \centering
        \includegraphics[width=\linewidth]{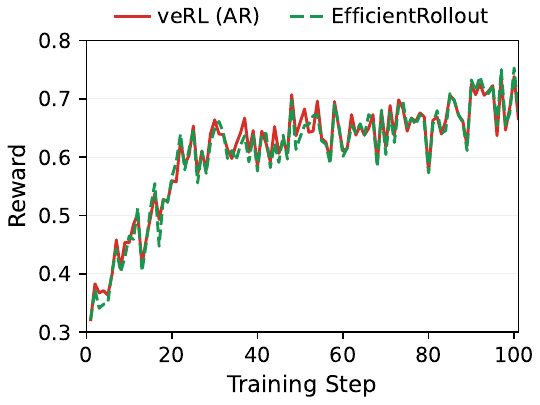}
        \caption{Qwen2.5-14B}
        \label{fig:appendix_qwen14b_reward}
    \end{subfigure}
    \hfill
    \begin{subfigure}[t]{0.32\linewidth}
        \centering
        \includegraphics[width=\linewidth]{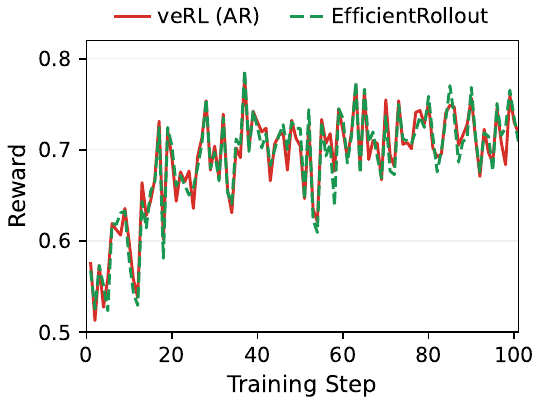}
        \caption{Llama3.1-8B}
        \label{fig:appendix_llama8b_reward}
    \end{subfigure}

    \caption{
    Average training reward over RL steps for three evaluated models.
    Across all evaluated models, \methodtitle{} closely follows the No-SD reward trajectory, suggesting that rollout acceleration preserves the main training dynamics.
    }
    \label{fig:app_fidelity_reward}
\end{figure}
\begin{figure}[t]
    \centering

    \begin{subfigure}[t]{0.32\linewidth}
        \centering
        \includegraphics[width=\linewidth]{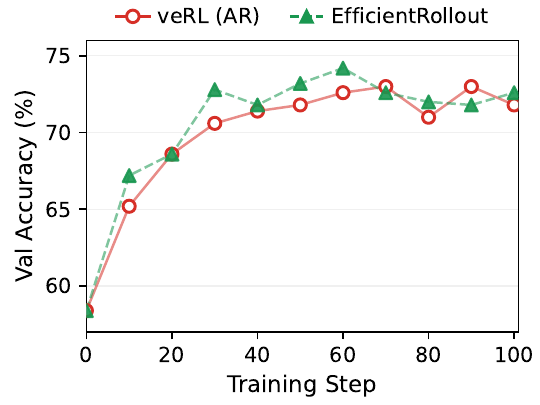}
        \caption{Qwen2.5-7B}
        \label{fig:app_qwen7b_valscore}
    \end{subfigure}
    \hfill
    \begin{subfigure}[t]{0.32\linewidth}
        \centering
        \includegraphics[width=\linewidth]{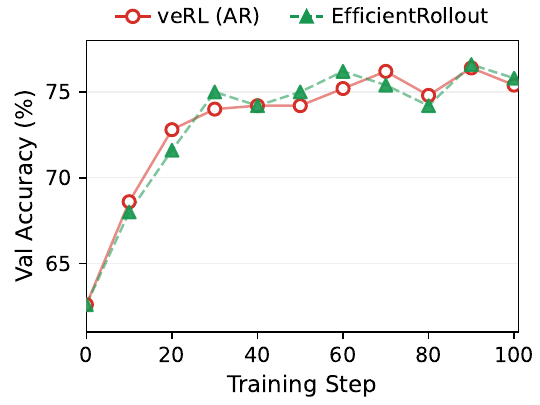}
        \caption{Qwen2.5-14B}
        \label{fig:app_qwen14b_valscore}
    \end{subfigure}
    \hfill
    \begin{subfigure}[t]{0.32\linewidth}
        \centering
        \includegraphics[width=\linewidth]{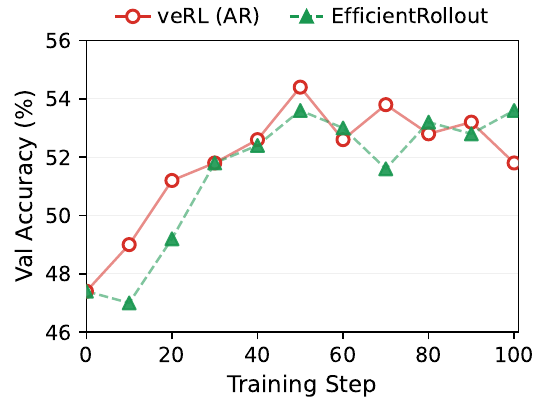}
        \caption{Llama3.1-8B}
        \label{fig:app_llama8b_valscore}
    \end{subfigure}

    \caption{
    Validation accuracy over training steps for the three evaluated models.
    \methodtitle{} achieves validation trajectories comparable to the \nosd{} baseline, supporting that the acceleration does not degrade downstream model quality in our evaluated settings.
    }
    \label{fig:app_fidelity_valscore}
\end{figure}

\section{Additional Evaluation Results}

\subsection{Per-Step Rollout Generation Time}
\label{app:gen_time_over_time}

\Cref{fig:app_gen_time_breakdown} shows rollout-generation time over training steps across models and SD baselines for RL rollouts.
\methodtitle{} achieves the lowest generation time throughout training on Qwen2.5-7B and Llama3.1-8B, and for most Qwen2.5-14B steps.
The small late-stage slowdown on Qwen2.5-14B coincides with longer sampled responses under \methodtitle{}, whose maximum response length reaches about 4,108 tokens, compared with 3,383 for \learned{} and 2,916 for \history{}.
We view this fluctuation as response-length stochasticity rather than a systematic toggle-policy failure.

\Alwayssd{} can improve over \nosd{} on Qwen2.5-7B and Llama3.1-8B, but can become slower when SD is enabled throughout rollout, especially on Qwen2.5-14B where the larger quantized drafter increases drafting cost.
\Learned{} achieves reasonable rollout speedup on Qwen2.5-14B because its auxiliary drafter combines low proposal cost with moderate block efficiency, as shown in~\cref{tab:eagle3_timing_decomp}.
These per-step trends are consistent with the aggregate timing results in~\cref{tab:e2e_speedup}.

\subsection{Training Dynamics and Quality Preservation}
\label{app:quality-preservation}

\Cref{fig:app_fidelity_reward,fig:app_fidelity_valscore} report training reward and validation accuracy trajectories over RL training.
Across Qwen2.5-7B, Qwen2.5-14B, and Llama3.1-8B-Instruct, \methodtitle{} closely follows the \nosd{} baseline in both metrics.
These results support that the rollout speedups do not materially alter training dynamics or degrade downstream model quality in our evaluated settings.

\paragraph{Repeated-run consistency.}
Across three paired data-order seeds for Qwen2.5-7B, \methodtitle{} reduces rollout-generation and full-step latency by $22.2\%\pm1.8$ pp and $14.0\%\pm1.5$ pp (mean$\pm$SD), respectively.
Final validation accuracy remains comparable between AR and \methodtitle{} ($73.2\%\pm1.6$ pp vs.\ $73.3\%\pm0.3$ pp), with no meaningful difference in late-training reward.

\begin{figure}[t]
    \centering
    \includegraphics[width=0.5\linewidth]{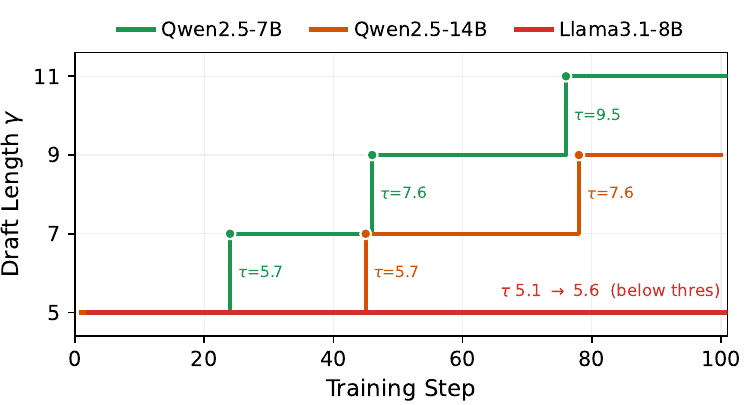}
    \caption{
    Adaptive $\gamma$ schedules across models. 
    The controller increases $\gamma$ only when block efficiency $\tau$ is high enough to support a longer draft; otherwise, it keeps $\gamma$ unchanged.
    }
    \label{fig:app_gamma_elevation}
\end{figure}

\FloatBarrier

\subsection{Adaptive Draft-Length Schedules Across Models}
\label{app:adaptive_gamma_schedule}

\Cref{fig:app_gamma_elevation} shows the $\gamma$ schedule selected by the adaptive controller across models.
Qwen2.5-7B increases from $5 \rightarrow 7 \rightarrow 9 \rightarrow 11$ as its \MAL{} repeatedly approaches the effective ceiling of the current draft length.
Before each elevation, its \MAL{} reaches $5.73$, $7.60$, and $9.48$, and after increasing $\gamma$ the \MAL{} further rises to $7.37$, $9.35$, and $11.25$, respectively.
Qwen2.5-14B follows the same pattern but later in training, increasing from $5 \rightarrow 7 \rightarrow 9$ when its \MAL{} reaches $5.73$ and $7.61$.
In contrast, Llama3.1-8B-Instruct stays at $\gamma=5$: although its \MAL{} improves from $5.07$ to around $5.5$, it never crosses the elevation threshold.
No downward adjustment is triggered in these runs.

\begin{table}[t]
\centering
\small
\setlength{\tabcolsep}{4.5pt}
\renewcommand{\arraystretch}{1.10}
\caption{Component-level decomposition of \methodtitle{} for Qwen2.5-7B over 100 steps. Each row adds one component to the preceding configuration using a common execution path.}
\label{tab:component_decomposition}
\vspace{4pt}
\begin{tabular}{lccc}
\toprule
\textbf{Sequential Configuration}
& \textbf{Rollout Gen. (s)}
& \textbf{Step Time (s)}
& \makecell{\textbf{Rollout / Step}\\\textbf{vs. AR (\%)}} \\
\midrule
\nosd{} & 82.4 & 132.6 & -- \\
$+$ Always-on W4 self-SD, fixed $\gamma=7$ & 75.6 & 127.6 & $-8.2 / -3.7$ \\
$+$ Roofline toggle, fixed $\gamma=7$ & 68.2 & 118.0 & $-17.2 / -11.0$ \\
\rowcolor{gray!10}
$+$ Adaptive $\gamma$ (full \methodtitle{}) & 66.3 & 115.7 & $-19.6 / -12.7$ \\
\bottomrule
\end{tabular}
\end{table}

\subsection{Component-Level Decomposition}
\label{app:component_decomposition}

\Cref{tab:component_decomposition} attributes 8.2/3.7, 9.0/7.3, and 2.4/1.7 pp of rollout/full-step reduction to W4 self-drafting, the Roofline toggle, and adaptive $\gamma$, respectively.
The absolute timings were measured before removing CPU-side bubbles from W4 drafting, whereas~\cref{tab:e2e_speedup} includes this optimization; because every row uses the same execution path, the factor-wise attribution remains controlled.

\begin{table}[t]
\centering
\small
\setlength{\tabcolsep}{5.0pt}
\renewcommand{\arraystretch}{1.10}
\caption{
Active-window analysis of \history{} after rollout history becomes available.
Statistics are reported from the second epoch onward.
}
\vspace{1mm}
\label{tab:history_based_slow}
\resizebox{0.85\linewidth}{!}{%
\begin{tabular}{lccccc}
\toprule
\textbf{Model}
&
\textbf{Resp. Len.}
&
\textbf{Reused Prefix}
&
\textbf{Reuse Rate}
&
\textbf{Verify Overhead}
&
\textbf{Gen. Time $\Delta$} \\
&
\textbf{(tokens)}
&
\textbf{(tokens)}
&
\textbf{(\%)}
&
\textbf{(s/step)}
&
\textbf{(s vs. No-SD)} \\
\midrule
Qwen2.5-7B
& 661.4
& 28.9
& 4.4
& 10.7
& +12.8 \\
Qwen2.5-14B
& 622.9
& 42.3
& 6.8
& 19.5
& +5.8 \\
Llama3.1-8B
& 646.2
& 324.6
& 50.2
& 15.9
& +11.1 \\
\bottomrule
\end{tabular}%
}
\end{table}

\subsection{Why History-Based Drafting Does Not Accelerate}
\label{app:why-history-slow}

\Cref{tab:history_based_slow} analyzes the rollout-history-based baseline after rollout history becomes available, \textit{i.e.}, from the second epoch onward.
This isolates the active window where history-based drafting can actually be applied.
Unlike fixed-$\gamma$ chain drafting, the \history{} method reuses variable-length prefixes from previous rollouts, so block efficiency $\mal$ and effective acceptance rate $\alpha$ are not directly comparable.
We therefore analyze prefix reuse, verification overhead, and rollout-generation time in this window.
Even in this favorable window, the method does not reduce rollout-generation time.

The main reason is that history reuse does not provide enough accepted tokens to amortize verification cost.
For Qwen2.5-7B, the Spec-RL implementation reuses only 30 tokens on average out of a 660-token response, corresponding to a 4.5\% reuse rate.
For Llama3.1-8B-Instruct, the reused prefix is longer, but verification is also more expensive, adding 15.9\,s of overhead per active step.
As a result, even after excluding the first epoch and using the original lossy lenience factor to increase acceptance, rollout-history-based drafting increases rollout generation time by 19.4\% on Qwen2.5-7B and 8.8\% on Llama3.1-8B-Instruct.
These results show that the history-based approach is limited not only by its cold start, but also by low effective reuse relative to verification cost.

\begin{figure}[t]
    \centering

    \begin{subfigure}[t]{0.32\linewidth}
        \centering
        \includegraphics[width=\linewidth]{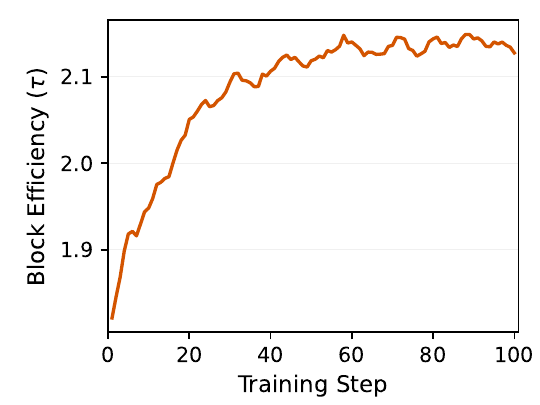}
        \caption{Qwen2.5-7B}
        \label{fig:appendix_qwen7b_eagle3_mal}
    \end{subfigure}
    \hfill
    \begin{subfigure}[t]{0.32\linewidth}
        \centering
        \includegraphics[width=\linewidth]{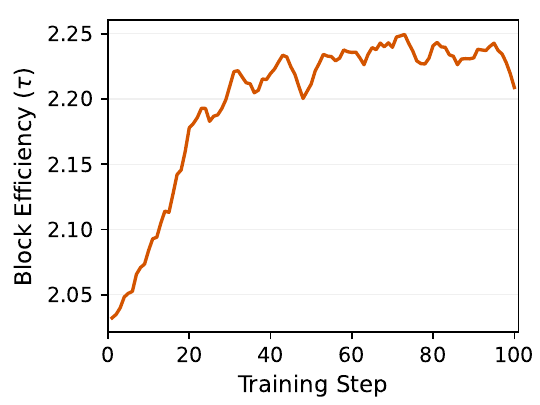}
        \caption{Qwen2.5-14B}
        \label{fig:appendix_qwen14b_eagle3_mal}
    \end{subfigure}
    \hfill
    \begin{subfigure}[t]{0.32\linewidth}
        \centering
        \includegraphics[width=\linewidth]{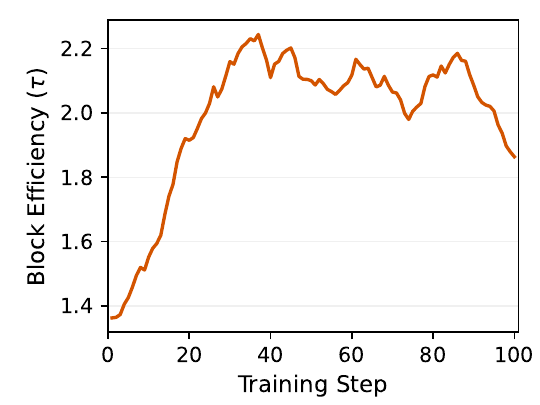}
        \caption{Llama3.1-8B}
        \label{fig:appendix_llama8b_eagle3_mal}
    \end{subfigure}

    \caption{
    Block efficiency of \learned{} over RL training steps. Across models, the pretrained drafter starts with low block efficiency but improves as online adaptation proceeds.
    }
    \label{fig:app_eagle3_mal}
\end{figure}
\begin{table}[t]
\centering
\scriptsize
\setlength{\tabcolsep}{4pt}
\renewcommand{\arraystretch}{1.05}
\caption{
Timing decomposition for \learned{}.
$\TargetTime$ is the single-token target forward time, $\DraftTime$ is the one-token drafter forward time, and $\VerifyTime$ is the target verification time for $\gamma+1$ tokens.
}
\label{tab:eagle3_timing_decomp}
\vspace{1mm}
\resizebox{0.65\linewidth}{!}{%
\begin{tabular}{lccccc}
\toprule
\textbf{Model}
&
$\TargetTime$ \textbf{(ms)}
&
$\DraftTime$ \textbf{(ms)}
&
$\VerifyTime$ \textbf{(ms)}
&
$\mathbf{\DraftTime/\TargetTime}$
&
$\mathbf{\VerifyTime/\TargetTime}$ \\
\midrule
Qwen2.5-7B
&
10.45
&
0.82
&
14.27
&
0.078
&
1.37 \\
Llama3.1-8B
&
13.66
&
0.85
&
19.75
&
0.062
&
1.45 \\
Qwen2.5-14B
&
20.31
&
0.91
&
28.50
&
0.045
&
1.40 \\
\bottomrule
\end{tabular}%
}
\end{table}

\begin{table}[t]
\centering
\scriptsize
\setlength{\tabcolsep}{4pt}
\renewcommand{\arraystretch}{1.05}
\caption{
Depth-dependent block efficiency for EAGLE3-based \learned{} before online adaptation.
Llama3.1-8B-Instruct already drops close to one after 512 generated tokens, while the Qwen models degrade more gradually.
}
\label{tab:eagle3_depth_tau_step1}
\vspace{1mm}
\resizebox{0.55\linewidth}{!}{%
\begin{tabular}{lccccc}
\toprule
\textbf{Model}
&
\textbf{0--512}
&
\textbf{512--1k}
&
\textbf{1k--2k}
&
\textbf{2k--4k}
&
\textbf{4k--8k} \\
\midrule
Qwen2.5-7B
&
1.939
&
1.895
&
1.723
&
1.526
&
1.195 \\
Llama3.1-8B
&
\textbf{2.252}
&
\textbf{1.418}
&
\textbf{1.100}
&
\textbf{1.037}
&
\textbf{1.056} \\
Qwen2.5-14B
&
2.045
&
2.037
&
2.029
&
2.237
&
1.866 \\
\bottomrule
\end{tabular}%
}
\end{table}

\begin{table}[!t]
\centering
\scriptsize
\setlength{\tabcolsep}{4pt}
\renewcommand{\arraystretch}{1.05}
\caption{
Depth-dependent block efficiency for EAGLE3-based \learned{} over the first five RL steps.
Llama3.1-8B-Instruct exhibits a sharp drop after 512 generated tokens, whereas the Qwen models maintain or recover block efficiency at deeper positions.
}
\label{tab:eagle3_depth_mal}
\vspace{1mm}
\resizebox{0.55\linewidth}{!}{%
\begin{tabular}{lccccc}
\toprule
\textbf{Model}
&
\textbf{0--512}
&
\textbf{512--1k}
&
\textbf{1k--2k}
&
\textbf{2k--4k}
&
\textbf{4k--8k} \\
\midrule
Qwen2.5-7B
&
1.970
&
1.921
&
1.790
&
1.716
&
1.604 \\
Llama3.1-8B
&
\textbf{2.288}
&
\textbf{1.557}
&
\textbf{1.252}
&
\textbf{1.190}
&
\textbf{1.294} \\
Qwen2.5-14B
&
2.070
&
2.061
&
2.103
&
2.390
&
2.573 \\
\bottomrule
\end{tabular}%
}
\end{table}

\subsection{Why Learned Auxiliary Drafting Remains Challenging}
\label{app:why-learned-auxiliary-challenging}

\paragraph{Block efficiency of \learned{}.}

EAGLE3-style auxiliary drafters typically require pretraining and online adaptation to match the target model and rollout data distribution~\citep{zhang2025fastgrpo, hu2026taming, iso2026accelerating}.
In our setting, a randomly initialized drafter yields \MAL{} close to 1.0, meaning that it rarely predicts even the first token correctly.
As shown in \cref{fig:app_eagle3_mal}, starting from a generally pretrained drafter improves early-stage behavior, but the initial \MAL{} remains limited: with a ShareGPT-pretrained drafter evaluated on SimpleRL-math rollouts, \MAL{} starts around 1.4--2.0.
A drafter pretrained directly on the target RL post-training distribution would likely provide higher initial \MAL{}~\citep{iso2026accelerating}; however, such in-distribution drafter pretraining is not always available in practice, so we use a drafter pretrained on general chat/text data~\citep{li2025eagle}.

\paragraph{Limited acceleration in Qwen-7B/14B.}

For the Qwen2.5-series models, learned auxiliary drafting provides limited generation-time acceleration mainly because its \MAL{} remains insufficient for our RL rollout workload.
Under high-temperature sampling ($T=1.0$) and long reasoning generation, the pretrained drafters do not provide sufficient proposal quality for our math RL rollout distribution.
In addition, system-aware activation remains important: as indicated by the gap between \alwayssd{} and \methodtitle{}, enabling SD during early large-batch phases can reduce or eliminate its benefit.
Higher \MAL{} may be achievable with in-distribution drafter initialization or more aggressive online adaptation, such as training the auxiliary drafter for multiple steps per RL iteration or running adaptation asynchronously.
However, these choices introduce additional drafter-training, scheduling, and configuration search burden; we further analyze the workload-matching requirement in~\cref{app:workload-matched-auxiliary-drafters}.

\paragraph{Unexpected slowdown on Llama3.1-8B.}

\citet{li2025eagle} report strong inference speedups for EAGLE3 in standard serving settings across Qwen and Llama models.
However, these results are not directly comparable to our RL rollout setting, which combines vLLM execution, high-temperature sampling ($T=1.0$), hard reasoning prompts, and long response budgets.
To diagnose the Llama3.1-8B-Instruct slowdown, we run short cold-start profiling experiments for the first five RL steps under the same rollout setting as in \cref{sec:eval-setting}, using vLLM EAGLE3 timing (\texttt{VLLM\_SD\_TIMING=1}), position-binned \MAL{} logging, and a custom proposal timer.

\Cref{tab:eagle3_timing_decomp} shows that the slowdown is not primarily due to drafter cost: the one-token drafter time $\DraftTime$ is similar across models, around 0.8--0.9\,ms.
Llama3.1-8B-Instruct has the largest verification-to-target ratio ($\VerifyTime/\TargetTime=1.45$), but this is only modestly higher than the Qwen models.
Thus, timing overhead alone does not explain the much larger slowdown.

The larger difference appears in depth-dependent block efficiency.
As shown in~\cref{tab:eagle3_depth_tau_step1}, before online adaptation, Llama3.1-8B-Instruct already shows block efficiency close to one after 512 generated tokens, indicating weak long-depth proposal quality from the pretrained drafter.
After the first five online-adaptation steps, this weakness persists: Llama3.1-8B-Instruct remains near 1.2--1.3 in the 1k--8k bins, whereas the Qwen models quickly recover block efficiency relative to their pre-adaptation values, especially at deeper positions (\cref{tab:eagle3_depth_tau_step1,tab:eagle3_depth_mal}).
This is precisely the regime that matters for rollout latency.
Long-tail responses dominate the rollout makespan, and the Llama drafter pays draft and verification overhead while most proposals are rejected, making the long tail even more expensive.
This suggests a model- and workload-dependent failure mode of learned auxiliary drafting rather than a stable speedup across RL rollouts.

\begin{figure}[t]
    \centering
    \includegraphics[width=0.7\linewidth]{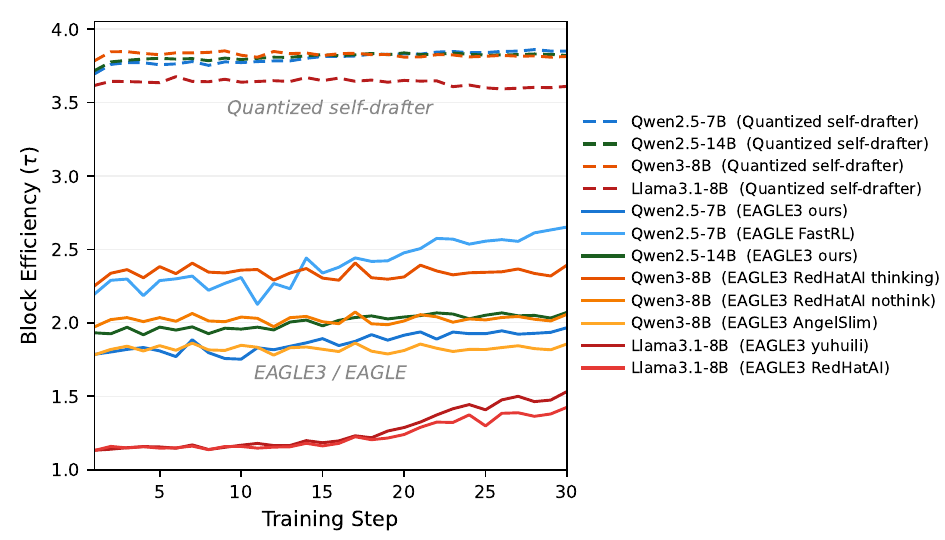}
    \caption{
    Block efficiency on DAPO-Math-17K.
    Across the first 30 RL steps, evaluated learned auxiliary drafters remain largely below target-induced quantized drafters.
    }
    \label{fig:app_eagle3_dapo_mal}
\end{figure}

\begin{figure}[t]
    \centering
    \begin{subfigure}[t]{0.32\linewidth}
        \centering
        \includegraphics[width=\linewidth]{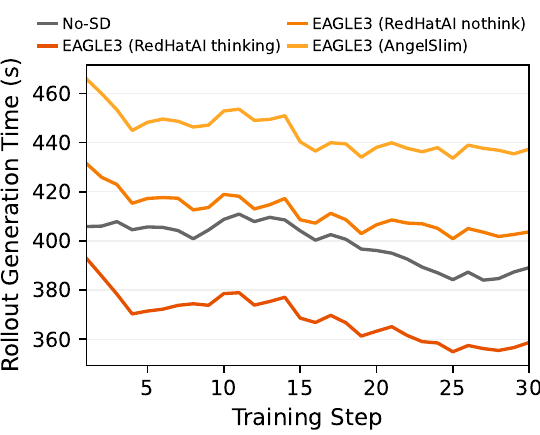}
        \caption{Qwen3-8B}
        \label{fig:appendix_qwen8b_eagle3_gentime}
    \end{subfigure}
    \hfill
    \begin{subfigure}[t]{0.32\linewidth}
        \centering
        \includegraphics[width=\linewidth]{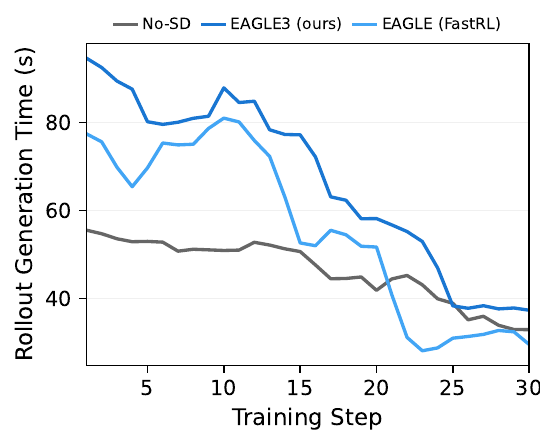}
        \caption{Qwen2.5-7B}
        \label{fig:appendix_qwen7b_eagle3_gentime}
    \end{subfigure}
    \hfill
    \begin{subfigure}[t]{0.32\linewidth}
        \centering
        \includegraphics[width=\linewidth]{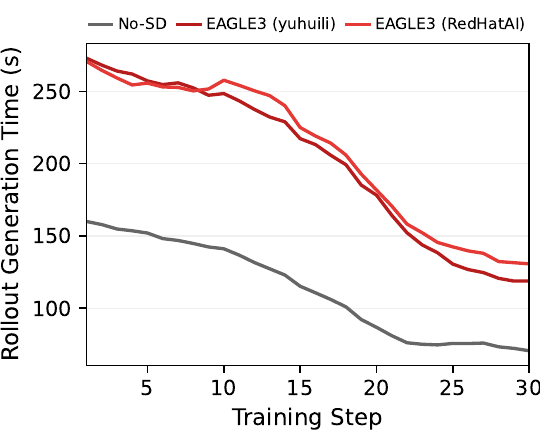}
        \caption{Llama3.1-8B}
        \label{fig:appendix_llama8b_eagle3_gentime}
    \end{subfigure}

    \caption{
    Rollout-generation time of EAGLE3 in the NeMo RL stack.
    Among the evaluated EAGLE3 drafters, only Qwen3-8B with the RedHatAI thinking drafter reduces rollout-generation time, while the other configurations are slower than No-SD due to insufficient block efficiency.
    }
    \label{fig:app_eagle3_dapo_gentime}
    \vspace{-2pt}
\end{figure}

\subsection{Auxiliary Drafters Aligned with RL Rollout Distributions Are Difficult to Obtain}
\label{app:workload-matched-auxiliary-drafters}

\paragraph{NeMo-RL-native validation setup.}

To check whether the moderate \MAL{} of \learned{} stems from our veRL~\citep{sheng2025hybridflow}-based implementation or from the drafter itself, we additionally evaluate EAGLE3 drafters in NVIDIA's native NeMo RL SD stack~\citep{iso2026accelerating}.
This stack uses Megatron~\citep{shoeybi2019megatron}-based online drafter training with vLLM rollout serving, and we leave the core RL, SD, and reward paths unchanged.
We train on the DAPO-Math-17K dataset~\citep{yu2025dapo} using one node with 8 H100-SXM GPUs, $\DraftLength=3$ with chain drafting and verification, 30 RL steps, a maximum generation length of 8k, temperature $T=1.0$, and a global batch size of 1k rollouts.
For comparison, we also measure the \MAL{} of quantized self-drafters under the same DAPO-Math experimental setting in our self-SD stack.

\paragraph{Evaluated learned auxiliary drafters.}

For Qwen3-8B, we evaluate three public EAGLE3 drafters: RedHatAI's thinking drafter~\citep{redhatai2025qwen3eagle3thinking}, RedHatAI's non-thinking drafter~\citep{redhatai2025qwen3eagle3nothink}, and AngelSlim's drafter~\citep{angelslim2025qwen3eagle3}.
For Qwen2.5-7B and Qwen2.5-14B, we did not find compatible public EAGLE3 checkpoints, so we use our ShareGPT-pretrained drafters introduced in~\cref{app:learned_auxiliary_drafting}.
For Llama3.1-8B-Instruct, we evaluate both the RedHatAI public drafter~\citep{redhatai2025llama31eagle3} and the official checkpoint from the EAGLE3 authors released by yuhuili~\citep{yuhuili2024llama31eagle3}.
We also evaluate the public Qwen2.5-7B EAGLE-style drafter~\citep{mithanlab2026qwen25eagle} released with FastRL~\citep{hu2026taming} as an RL-targeted auxiliary-drafter checkpoint.
The checkpoint is reported to be trained on the OpenThoughts2-1M dataset~\citep{openthoughts2026openthoughts2}, a synthetic reasoning dataset spanning math, science, code, and puzzles.
Because this checkpoint is not directly compatible with NeMo RL's EAGLE3-specific online update path, we use it as a fixed drafter over the 30 RL steps without online adaptation.

\paragraph{Block efficiency gap.}

\Cref{fig:app_eagle3_dapo_mal} shows that evaluated learned auxiliary drafters achieve substantially lower block efficiency than target-induced quantized drafters.
Across the first 30 RL steps, quantized self-drafters stay near $\mal{}=3.6$--$3.9$ under $\DraftLength=3$, close to the ceiling of 4.
By contrast, EAGLE3 drafters remain mostly in the $\mal{}=1.2$--$2.4$ range, with the best public configuration, Qwen3-8B with the RedHatAI thinking drafter, reaching roughly $\mal{}=2.2$--$2.4$.
The FastRL Qwen2.5-7B EAGLE-style drafter reaches higher \MAL{} ($\mal{}=2.2$--$2.6$) than our ShareGPT-pretrained EAGLE3 drafter, but this is still insufficient for consistent rollout-generation speedup.
Within our 30-step window, online adaptation yields only modest \MAL{} improvement from the evaluated initializations.
These results suggest that the limitation is not the EAGLE/EAGLE3 architecture itself, but the availability of an auxiliary drafter sufficiently aligned with long, high-temperature RL rollout distributions and effective from the beginning of RL training.

\paragraph{Generation-time consequence.}

\Cref{fig:app_eagle3_dapo_gentime} shows the practical consequence of low block efficiency within the NeMo RL stack.
All timing curves compare \learned{} against the corresponding NeMo RL No-SD baseline.
Among the evaluated EAGLE3 drafters, Qwen3-8B with the RedHatAI thinking drafter is the only configuration that consistently reduces rollout-generation time, improving it by about 7.7\%.
The Qwen2.5-7B EAGLE-style FastRL drafter becomes consistently beneficial only late in training, reducing rollout-generation time by 17.6\% over steps 22--30.
The other configurations are slower than No-SD for much of training because their \MAL{} is too low to amortize draft, verification, and online-update overhead.
This confirms that learned auxiliary drafting can reduce rollout-generation time when the drafter reaches sufficiently high \MAL{}.

\paragraph{Output length and training distribution.}

One possible explanation is an output-sequence-length mismatch between drafter training and rollout inference.
However, when we run ablation probes with shorter output caps (e.g., 1k, 2k, and 4k), most public checkpoints or drafters pretrained on general chat/text data remain weak even under shorter caps, and several configurations are roughly length-flat.
This suggests that output length alone does not explain their low block efficiency.
A more central factor appears to be the drafter-training distribution: public drafters and our Qwen2.5 drafters are trained on general chat/text corpora such as ShareGPT or UltraChat, rather than target-generated rollouts from the RL workload.
\citet{hu2026taming} train their drafter on OpenThoughts2-1M~\citep{openthoughts2026openthoughts2}, a synthetic reasoning dataset closer to long reasoning workloads than general chat corpora.
More generally, reaching the high-\MAL{} regime reported by~\citet{iso2026accelerating} may require a more specialized auxiliary-drafter training pipeline, such as collecting target-generated long rollouts from the RL workload and then training the auxiliary drafter on the resulting data.

\paragraph{Takeaway.}

Our results show that reaching this regime depends on a well-aligned auxiliary drafter, which is not generally available from public checkpoints or generic drafter-training recipes.
Obtaining such a drafter can require a separate pipeline for target-generated long-rollout collection, offline auxiliary-drafter training, and possibly more aggressive online adaptation.
This raises the barrier to using \learned{} as a drop-in rollout accelerator.
Target-induced self-drafting provides a simpler point in this trade-off: it requires no external drafter checkpoint or offline auxiliary-drafter pretraining, yet achieves high \MAL{} from the beginning of RL training and translates it into practical rollout speedup.

\begin{table}[t]
\centering
\small
\setlength{\tabcolsep}{6pt}
\renewcommand{\arraystretch}{1.08}
\caption{Training results on H100 over 100 steps. We report \methodtitle{} acceptance rate and AR$\rightarrow$\methodtitle{} latency; signed percentages are relative to AR.}
\label{tab:h100_e2e}
\vspace{4pt}
\begin{tabular}{lccc}
\toprule
\textbf{Model}
& \textbf{$\alpha$ (\%) $\uparrow$}
& \textbf{Rollout Gen. (s) $\downarrow$}
& \textbf{Step Time (s) $\downarrow$} \\
\midrule
Qwen2.5-7B  & 98.2 & 60.1 $\rightarrow$ 50.1 (-16.7\%) & 86.5 $\rightarrow$ 76.3 (-11.8\%) \\
Qwen2.5-14B & 97.6 & 90.1 $\rightarrow$ 80.9 (-10.3\%) & 139.0 $\rightarrow$ 129.2 (-7.1\%) \\
Llama3.1-8B & 96.2 & 87.0 $\rightarrow$ 72.6 (-16.6\%) & 120.9 $\rightarrow$ 105.9 (-12.5\%) \\
\bottomrule
\end{tabular}
\end{table}

\begin{table}[t]
\centering
\small
\setlength{\tabcolsep}{4.5pt}
\renewcommand{\arraystretch}{1.08}
\caption{End-to-end evaluation for Qwen2.5-32B with TP$=2$ on A100 over 100 steps. Metrics follow~\cref{tab:e2e_speedup}; signed percentages are relative to AR.}
\label{tab:qwen32_tp2_e2e}
\vspace{4pt}
\begin{tabular}{lccccc}
\toprule
\textbf{Drafting Method}
& \textbf{Prep. (s) $\downarrow$}
& \textbf{$\mal$ / $\bar{\gamma}$}
& \textbf{$\alpha$ (\%) $\uparrow$}
& \textbf{Rollout Gen. (s) $\downarrow$}
& \textbf{Step Time (s) $\downarrow$} \\
\midrule
AR & -- & -- & -- & 162.4 & 308.7 \\
\rowcolor{gray!10}
\methodtitle{} & 3.3 & 9.0 / 8.6 & 98.5 & 141.2 (-13.0\%) & 289.0 (-6.4\%) \\
\bottomrule
\end{tabular}
\end{table}

\begin{table}[t!]
\centering
\small
\setlength{\tabcolsep}{5pt}
\renewcommand{\arraystretch}{1.08}
\caption{Additional matched 50-step Qwen2.5-7B evaluations on A100. Response lengths are in tokens; Avg. Max Response is the mean length of the longest sampled response across steps.}
\label{tab:workload_e2e}
\vspace{4pt}
\begin{tabular}{ccccc}
\toprule
\textbf{Temperature}
& \textbf{Max Response}
& \textbf{Avg. Max Response}
& \textbf{$\alpha$ (\%)}
& \textbf{Rollout / Step Reduction (\%)} \\
\midrule
0.6 & 8,192  & 8,157  & 99.2 & 27.2 / 19.0 \\
1.0 & 16,384 & 7,363  & 97.6 & 24.3 / 18.6 \\
0.6 & 16,384 & 13,561 & 99.0 & \textbf{40.1 / 34.0} \\
\bottomrule
\end{tabular}
\end{table}

\subsection{Extended Evaluation Across Hardware, TP, and Workload Settings}
\label{app:extended_e2e_evaluation}

\paragraph{Cross-hardware H100 evaluation.}
We evaluate on H100 using the main evaluation configuration, recalibrating once per model--hardware pair.
\Cref{tab:h100_e2e} shows that \methodtitle{} maintains a 96.2--98.2\% acceptance rate while reducing rollout-generation and full-step latency by 10.3--16.7\% and 7.1--12.5\%, respectively.

\paragraph{Capacity-driven TP$=2$ execution.}
\methodtitle{} naturally supports end-to-end execution with tensor parallelism (TP) by incorporating the calibrated collective term in~\cref{eq:tp_aware_roofline} into its toggle policy.
\Cref{tab:qwen32_tp2_e2e} shows that, for Qwen2.5-32B with capacity-driven TP$=2$ on 8$\times$A100 GPUs, \methodtitle{} reduces rollout-generation and full-step latency by 13.0\% and 6.4\%, respectively.
In TP settings, irreducible collective costs raise the effective $\DraftTime/\TargetTime$ ratio and narrow the SD-beneficial region.
When the target and drafter fit at TP$=1$, data-parallel rollout workers therefore provide a more favorable latency regime.

\paragraph{Temperature and response-length settings.}
We additionally evaluate lower-temperature and 16k settings that reflect prior reasoning-RL practice: DAS~\citep{shao2025beat} uses a temperature of $0.6$ with a 16k maximum sequence length, while DeepSeek-R1~\citep{guo2025deepseek} lowers its temperature to $0.7$ in the second RL stage.
According to~\cref{tab:workload_e2e}, increasing the maximum response length makes the realized response distribution more long-tailed: at temperature $0.6$, the average per-step maximum response nearly saturates the 8k response-length cap (8,157 tokens) and rises to 13,561 tokens under the 16k cap, while rollout-generation/full-step reductions increase from 27.2/19.0\% to 40.1/34.0\%.
At a fixed 16k cap, lowering temperature from 1.0 to 0.6 raises $\alpha$ from 97.6\% to 99.0\% and increases the reductions from 24.3/18.6\% to 40.1/34.0\%, consistent with a prior study~\citep{ouyang2024temperature} showing that lower temperatures improve drafter acceptance rate and speedup.



\end{document}